\documentclass[12pt]{article}
\usepackage{enumitem}
\usepackage{fullpage}
\usepackage[english]{babel}
\usepackage[utf8]{inputenc}
\usepackage[authoryear,comma]{natbib}
\usepackage{amsmath,multibib}
\usepackage{bbm}
\usepackage{amssymb,amsthm,mathabx,mathtools}
 \usepackage{relsize}
\usepackage{mathrsfs}
\usepackage{graphicx}
\usepackage{xurl}
\usepackage{caption,subcaption}
\usepackage{booktabs}
\usepackage{multirow}
\usepackage{caption}
\usepackage{epsfig,float}
\usepackage{url} 
\usepackage[usenames,dvipsnames]{color} 
\usepackage{listings}
\usepackage{mathrsfs,bm} 

\usepackage{tikz}
\usepackage{graphicx}

\theoremstyle{plain}
\newtheorem{thm}{Theorem}

\newcommand{\bthm}{\begin{thm}}
\newcommand{\ethm}{\end{thm}}
\newcommand{\bpf}{\begin{proof}}
\newcommand{\epf}{\end{proof}}
\theoremstyle{definition}
\newtheorem{defn}{Definition}
\newtheorem{example}{Example}
\newtheorem{rem}{Remark}
\usepackage[margin=.9in,a4paper]{geometry}
\usepackage[compact]{titlesec}
\numberwithin{equation}{section}
\usepackage{deep-macro}
\usepackage{epigraph}
\usepackage{relsize}

\newcommand{\bal}{\mathlarger{\alpha}}
\newcommand{\hbal}{\mathlarger{\widehat\alpha}}

\def\ci{\perp\!\!\!\perp}

\newcommand{\sbt}{\,\begin{picture}(-1,1)(-1,-3)\circle*{3}\end{picture}~\ }

\usepackage{relsize}
\usepackage[most]{tcolorbox}
\usepackage{tikz}
\usetikzlibrary{automata, positioning}
\usetikzlibrary{shapes.geometric}
\usepackage{bbm}
\usepackage{tabularx}
\newcolumntype{Y}{>{\centering\arraybackslash}X}
\newcolumntype{h}{>{\hsize=.5\hsize\centering\arraybackslash\extracolsep{.1em}}X}
\newcolumntype{q}{>{\hsize=.75\hsize\centering\arraybackslash\extracolsep{.1em}}X}

\begin{document}

\begin{center}
{\bf 
{\Large InfoGram and Admissible Machine Learning}\\[1.1em]

Deep Mukhopadhyay}\\[.1em]
\texttt{deep@unitedstatalgo.com}\\[.3em]

\end{center}
\linespread{1.35}
\renewcommand{\baselinestretch}{1.35}
\setlength{\parskip}{1ex}

\vspace{.2em}
\begin{abstract} 
We have entered a new era of machine learning (ML), where the most accurate algorithm with superior predictive power may not even be deployable, unless it is \textit{admissible} under the regulatory constraints. This has led to great interest in developing fair, transparent and trustworthy ML methods. The purpose of this article is to introduce a new information-theoretic learning framework (admissible machine learning) and algorithmic risk-management tools (InfoGram, L-features, ALFA-testing) that can guide an analyst to \textit{redesign} off-the-shelf ML methods to be regulatory compliant, while maintaining good prediction accuracy. We have illustrated our approach using several real-data examples from financial sectors, biomedical research, marketing campaigns, and the criminal justice system.

\end{abstract} 
\vspace{-.3em}
\noindent\textsc{\textbf{Keywords}}: 
Admissible machine learning; InfoGram; L-Features; Information-theory; ALFA-testing, Algorithmic risk management; Fairness; Interpretability;  COREml; FINEml. 

\vspace{-.12em}
\renewcommand{\baselinestretch}{.18}
\setlength{\parskip}{0ex}
\renewcommand\contentsname{}
{\small
\setcounter{tocdepth}{3}
\tableofcontents
}
\vskip.2em
\linespread{1.29}
\renewcommand{\baselinestretch}{1.3}
\setlength{\parskip}{1.5ex}
\vskip1.7em
\subsection*{Category: Fairness, Explainability, and Algorithm Bias}
\fbox{
\begin{minipage}{37em}
\vskip.64em
Machine learning (ML) methods are rapidly becoming an essential part of automated decision-making systems that directly affect human lives. While substantial progress has been made toward developing more powerful computational algorithms, the widespread adoption of these technologies still faces several barriers---the biggest one being
ensuring adherence to regulatory requirements, without compromising too much accuracy. Naturally, the question arises: how to systematically go about building such regulatory-compliant fair and trustworthy algorithms? This paper offers new statistical principles and information-theoretic graphical exploratory tools that engineers can use to ``detect, mitigate, and remediate'' off-the-shelf ML-algorithms, thereby making them \textit{admissible} under appropriate laws and regulatory scrutiny.
\vskip.24em
\end{minipage}
}

\section{Introduction}
First-generation ``prediction-only'' machine learning technology has served the tech and eCommerce industry pretty well. However, ML is now rapidly expanding beyond its traditional domains into highly regulated or safety-critical areas---such as healthcare, criminal justice systems,  transportation, financial markets, and national security---where achieving high predictive-accuracy is often as important as ensuring regulatory compliance and transparency in order to ensure the trustworthiness. We thus focus on developing \textit{admissible machine learning} technology that can balance fairness, interpretability, and accuracy in the best manner possible. How to systematically go about building such algorithms in a fast and scalable manner?  This article introduces some new statistical learning theory and information-theoretic graphical exploratory tools to address this question.
\vskip.35em
{\bf Going Beyond ``Pure'' Prediction Algorithms}. Predictive accuracy is not the be-all and end-all  for judging the `quality' of a machine learning model. Here is a dazzling example: Researchers at the Icahn School of Medicine at Mount Sinai in New York City found that \citep{zech2018variable,reardon2019rise} a deep-learning algorithm, which showed more than 90\% accuracy on the x-rays produced at Mount Sinai, performed poorly when tested on data from other institutions. Later it was found that ``the algorithm was also factoring in the odds of a positive finding based on how common pneumonia was at each institution—not something they expected or wanted.'' This sort of unreliable and inconsistent performance can be clearly dangerous. As a result of these safety concerns, despite lots of hype and hysteria around AI in imaging, only about 30\% of radiologists are currently using machine learning (ML) for their everyday clinical practices \citep{allen2021}. To apply machine learning appropriately and safely-- especially when human life is at stake--we have to think beyond predictive accuracy. The deployed algorithm needs to be comprehensible (by end-users like doctors, judges, regulators, researchers, etc.) in order to make sure it has learned \textit{relevant and admissible features} from the data, which is meaningful in light of investigators’ domain knowledge. The fact of the matter is, an algorithm that is solely focused on \textit{what} is learned, without reasoning \textit{how} it learned what it has learned, is not intelligent enough. We next expand on this issue using two real data applications. 
\vskip.35em
{\bf Admissible ML for Industry}. Consider the UCI Credit Card data (discussed in more details in Sec \ref{sec:FINAPP}), collected in October 2005, from an important Taiwan-based bank. We have records of $n=30,000$ cardholders. The data composed of a response variable $Y$ denoting: default payment status (Yes = 1, No = 0), along with $p=23$ predictor variables (e.g., gender, education, age, history of past payment, etc.).  The goal is to accurately predict the probability of default given the profile of a particular customer.  

On the surface, this seems to be a straightforward classification problem for which we have a large inventory of powerful algorithms. \citet{yeh2009comparisons} performed an exhaustive comparison between six machine learning methods (logistic regression, K-nearest neighbor, neural net, etc.) and finally selected the neural network model, which attained $83\%$ accuracy on a 80-20 train-test split of the data. However, traditionally build ML models are not deployable, unless it is \textit{admissible} under the financial regulatory constraints\footnote{The Equal Credit Opportunity Act (ECOA) is a major federal financial regulation law enacted in 1974.} \citep{wall2018some}, which demand that (i) the method should not discriminate people on the basis of protective features\footnote{\texttt{https://en.wikipedia.org/wiki/Protected\_group}}, here based on \texttt{gender} and \texttt{age}; and (ii) The method should be simpler to interpret and transparent (compared to those big neural-nets or ensemble models like random forest and gradient boosting). 
\begin{figure}[ ]
\vspace{-.7em}
    \centering
  \includegraphics[width=.42\linewidth,trim=1.25cm .5cm 1.25cm 0cm]{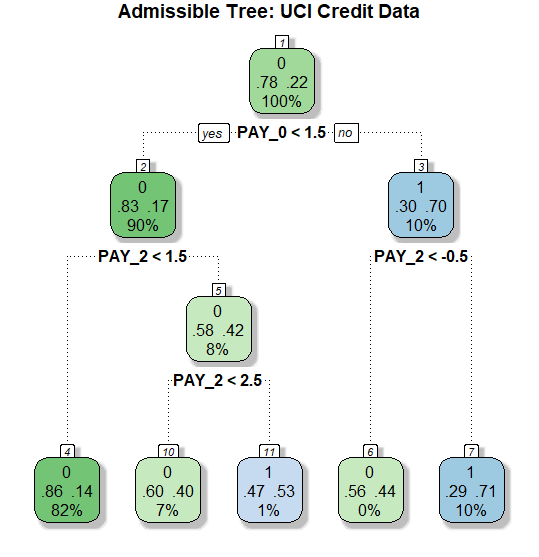}
  \vskip1em
\caption{A shallow admissible tree classifier for the UCI credit card data with four decision nodes, which is as accurate as the most complex state-of-the-art ML model.} \label{fig:intro}
  \end{figure}

\begin{figure}[ ]
\vspace{-.64em}
    \centering
  \includegraphics[width=.4\linewidth,trim=1.7cm .5cm 1.7cm 0cm]{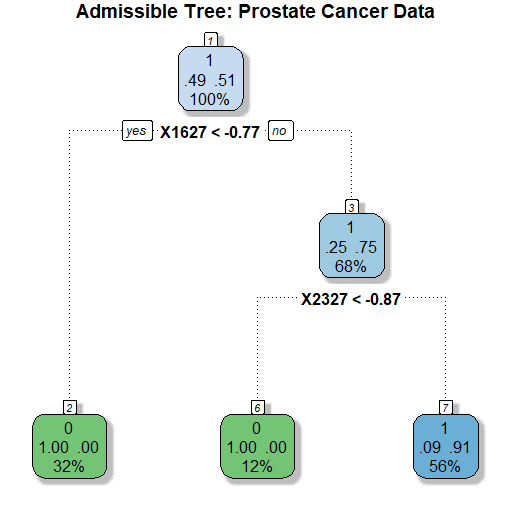}
  \vskip.5em
\caption{A two-gene admissible tree classifier for prostate cancer data with $p=6033$ gene expression measurements on $50$ control and $52$ cancer patients.} \label{fig:intro2}
  \end{figure}

To improve fairness, one may remove the sensitive variables and go back to business as usual by fitting the model on the rest of the features--known as `fairness through unawareness.' Obviously this is not going to work because there will be some proxy attributes (e.g, zip code or profession) that share some degree of correlation (information-sharing) with race, gender, or age. These proxy variables can then lead to the same unfair results. It is not clear how to define and detect those proxy variables to mitigate hidden biases in the data. In fact, on a recent review by \cite{roth2020} on algorithmic fairness, the authors forthrightly stated
\vspace{-.3em}
\begin{quote}
   {\small ``\textit{But despite the volume and
velocity of published work, our understanding
of the fundamental questions
related to fairness and machine
learning remain in its infancy.}''}
\end{quote}
\vspace{-.3em}
Currently, there exists no systematic method to directly construct an admissible algorithm that can mitigate bias. To quote a real practitioner of a reputed AI-industry: ``I ran 40,000 different random forest models with different features and hyper-parameters to search a fair model.'' This ad-hoc and inefficient strategy could be a significant barrier for an efficient large-scale implementation of admissible AI technologies. Fig. \ref{fig:intro} shows a fair and shallow tree classifier with four decision nodes, which attains 82.65\% accuracy; this was built in a completely automated manner without any hand-crafted manual tuning. Section \ref{sec:theory} will introduce the required theory and methods behind our procedure. Nevertheless,  this simple and transparent anatomy of the final model makes it easy to convey \textit{which} are the key drivers of the model:
variables \texttt{Pay\_0} and \texttt{Pay\_2}\footnote{\texttt{Pay\_0} and \texttt{Pay\_2} denote the repayment status of the last two months (-1=pay duly, 1=payment delay for one month, 2=payment delay for two months, and so on).} are the most important indicators to default. These variables have two key characteristics: they are highly predictive and at the same time safe to use in the sense that they share very little predictive information with the sensitive attributes age and gender, and for that reason, we call them \textit{admissible} features. The model also convey \textit{how} the key variables impacting credit risk: the simple decision tree shown in Fig. \ref{fig:intro} is fairly self-explanatory, and its clarity facilitates an easy explanation of the predictions. 
\vskip.35em
{\bf Admissible ML for Science}.  Legal requirement is not the only reason why we want to build admissible ML. In scientific investigations, it is important to know whether the deployed algorithm helps researchers to better understand the phenomena by refining their ``mental model.'' Consider, for example, the prostate cancer data where we have $p=6033$ gene expression measurements from 52 tumor and 50 normal specimens. Fig. \ref{fig:intro2} shows a 95\% accurate classification model for prostate data with only two ``core'' driver genes! This compact model is admissible in the sense that it confers the following benefits: (i) it identifies a two-gene signature (composed of gene-1627 and gene-2327) as the top factor associated with prostate cancer. They are \textit{jointly} overexpressed in the tumor samples but interestingly they have very little marginal information (not individually differentially expressed, as shown in Fig. \ref{fig:Proscart}). Accordingly, traditional linear-model-based analysis will fail to detect this gene-pair as a key biomarker.  (ii) The simple decision tree model in Fig. \ref{fig:intro2} provides a mechanistic understanding and justification as to why the algorithm thinks a patient has prostate cancer or not. (iii) Finally, it provides the needed guidance on what to do next by having a control over the system. In particular, a cancer biologist can choose between different diagnosis and treatment plans with the goal to regulate those two oncogenes.

\vskip.35em
{\bf  Goals and Organization}. The primary goal of this paper is to introduce some new fundamental concepts and tools to lay the foundation of \textit{admissible machine learning} that are efficient (enjoy good predictive accuracy), fair (prevent discrimination against minority groups), and interpretable (provide mechanistic understanding) to the best possible extent.

\vskip.25em

Our statistical learning framework is grounded in the foundational concepts of information theory. The required statistical formalism (nonparametric estimation and inference methods) and information-theoretic principles (entropy, conditional entropy, relative entropy, and conditional mutual information) are introduced in Section \ref{sec:theory}. 
A new nonparametric estimation technique for conditional mutual information (CMI) is proposed that scales to large datasets by leveraging the power of machine learning. For statistical inference, we have devised a new model-based bootstrap strategy. The method was applied to the problem of conditional independence testing and integrative genomics 
(breast cancer multi-omics data from Cancer Genome Atlas).  Based on this theoretical foundation, in Section \ref{sec:EAML}, we laid out the basic elements of admissible machine learning. Section \ref{sec:AINTER}
focuses on algorithmic interpretability:  how can we efficiently search and design self-explanatory algorithmic models by balancing accuracy and robustness to the best possible extent? Can we do it in a completely model-agnostic manner? Key concepts and tools introduced in this section are: Core features, infogram, L-features, net-predictive information, and COREml. The procedure was applied to several real datasets, including high-dimensional microarray gene expression datasets (prostate cancer and SRBCT data), MONK's problems, and Wisconsin breast cancer data. Section \ref{sec:AFAIR} focuses on algorithmic fairness, which tackles the challenging problem of designing admissible ML algorithms that are \textit{simultaneously} efficient, interpretable, and equitable. There are several key techniques introduced in this section: admissible feature selection, ALFA-testing, graphical risk assessment tool, and FINEml. We illustrate the proposed methods using examples from criminal justice system (ProPublica's COMPAS recidivism data), financial service industry (Adult income data, Taiwan credit card data), and marketing ad campaign. We conclude the paper in Section \ref{sec:con} by reviewing the challenges and opportunities of next-generation \textit{admissible} ML technologies.  
\section{Information-Theoretic Principles and Methods} \label{sec:theory}
The foundation of admissible machine learning relies on information-theoretic principles and nonparametric methods. The key theoretical ideas and results are presented in this section to develop a deeper understanding of the conceptual basis of our new framework.

\subsection{Notation} 
Let $Y$ be the response variable taking values $\{1,\ldots,k\}$, $\X=(X_1,\ldots,X_p)$ denotes a $p$-dimensional feature matrix, and   $\Sb=(S_1,\ldots,S_q)$ is additional set of $q$ covariates (e.g., collection of sensitive attributes like race, gender, age, etc.). A variable is called \texttt{mixed}  when it can take either discrete, continuous, or even categorical values, i.e., completely unrestricted data-types. Throughout, we will allow both $\X$ and $\Sb$ to be \texttt{mixed}. We write $Y \ci \X$ to denote the independence of $Y$ and $\X$. While, the conditional independence of $Y$ and $\X$ given $\Sb$ is denoted by $Y \ci \X \mid \Sb$. For a continuous random variable, $f$ and $F$ denote the probability density and distribution function, respectively. For a discrete random variable the probability mass function will be denoted by $p$ with proper subscript. 

\subsection{Conditional Mutual Information}
Our theory starts with an information-theoretic view of conditional dependence. Under conditional independence:
$$Y \ci \Xb \mid \Sb$$
the following decomposition holds for all $y,\xb,\sb$
\[ f_{Y,\X|\Sb}(y,\xb|\sb)\,=\, f_{Y|\Sb}(y|\sb) f_{\X|\Sb}(\xb|\sb).\]
More than testing independence, often the real interest lies in \textit{quantifying} the conditional dependence: the average deviation of the ratio
\beq \label{eq:ratio}
\dfrac{f_{Y,\X|\Sb}(y,\xb|\sb)}{f_{Y|\Sb}(y|\sb) f_{\X|\Sb}(\xb|\sb)},
\eeq
which can be measured by conditional mutual information \citep{wyner1978}.

\begin{defn}
Conditional mutual information (CMI) between $Y$ and $\X$ given $\Sb$ is defined as: 
\beq \label{eq:def}
\MI(Y,\X \mid \Sb)~=~\iiint_{y,\xb,\sb} \log \left( \dfrac{f_{Y,\X|\Sb}(y,\xb|\sb)}{f_{Y|\Sb}(y|\sb) f_{\X|\Sb}(\xb|\sb)}  \right) f_{Y,\X,\Sb}(y,\xb,\sb) \dd y \dd \xb \dd \sb.
\eeq
\end{defn}

{\bf Two Important Properties}. (P1) One of the striking features of CMI is that it captures multivariate non-linear conditional dependencies between the variables in a completely nonparametric manner. (P2) CMI possesses the necessary and sufficient condition as a measure of conditional independence, in the sense that 
\beq \MI(Y,\X|\Sb)=0 ~~\text{if and only if} ~~Y \ci \Xb \mid \Sb.\eeq

Conditional independence relation can be described using graphical model (also known as Markov network), as shown the figure below:
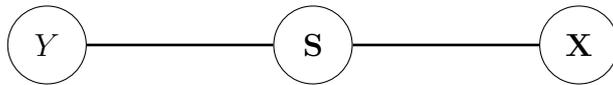
\begin{figure}[h]
\vspace{1em}
\centering
\begin{tikzpicture}[scale=1.4]
\node[state] at (0, 0) (y) {$Y$};
\node[state] at (2.5,0) (s) {$\Sb$};
\node[state] at (5, 0) (x) {$\X$};
\draw
(y) edge[line width=.35mm] node {} (s)
(s) edge[line width=.35mm] node {} (x);
\end{tikzpicture} 
\vskip.7em
\caption{Representing conditional independence graphically, where each node is a random variable (or random vector). The edge between $Y$ and $\X$ passes through the $\Sb$.} \label{fig:cig}
\vspace{-.83em}
\end{figure}

\subsection{Net-Predictive Information}
One of the major significances of CMI as a measure of conditional dependence comes from its interpretation in terms of additional `information gain' on $Y$ learned through $\Xb$ when we already know $\Sb$.  In other words, CMI measures the Net-Predictive Information (NPI) of $\boldmath{\Xb}$---the \textit{exclusive} information content of $\X$ for $Y$ beyond what is already subsumed by $\Sb$. To formally arrive at this interpretation, we have to look at CMI from a different angle, by expressing it in terms of conditional entropy. Entropy is a fundamental information-theoretic uncertainty measure. For a random variable $Z$, entropy $H(Z)$ is defined as $-\Ex_Z[\log f_Z]$.

\begin{defn}
The conditional entropy $H(Y|\Sb)$ is defined as the expected entropy of $Y|\Sb=\sb$
\beq 
H(Y\mid \Sb) = \int_\sb H(Y \mid \Sb=\sb) \dd F_\sb,
\eeq
which measures how much uncertainty remains in
$Y$ after knowing $\Sb$, on average. 
\end{defn}
 
\begin{thm} \label{thm:centropy}
For $Y$ discrete and $(\Xb,\Sb)$ \texttt{mixed} multidimensional random vectors, $\MI(Y,\Xb|\Sb)$ can be expressed as the difference between two conditional-entropy statistics:
\beq \label{eq:dce}
\MI(Y,\Xb \mid \Sb)\,=\,H(Y \mid \Sb)\,-\, H(Y \mid \Sb,\Xb).
\eeq
\end{thm}
The proof involves some standard algebraic manipulations, and is given in Appendix \ref{app:proof1}.

\begin{rem}[Uncertainty Reduction]
The alternative way of defining CMI through eq. \eqref{eq:dce} allows us to interpret it from a new angle: Conditional mutual information $\MI(Y,\Xb | \Sb)$ measures the \textit{net impact} of $\Xb$ in reducing  the uncertainty of $Y$, given $\Sb$. This new perspective will prove to be vital for our subsequent discussions. Note that, if $H(Y|\Sb,\Xb)=H(Y|\Sb)$, then $\Xb$ carries no \textit{net}-predictive information about $Y$.
\vspace{-.4em}
\end{rem}

\subsection{Nonparametric Estimation Algorithm}
The basic formula \eqref{eq:def} of conditional mutual information (CMI) that we have presented in the earlier section,  is, unfortunately, not readily applicable for two reasons. First, the practical side: in the current form, \eqref{eq:def} requires estimation of  $f_{Y,\X|\Sb}$ and $f_{\X|\Sb}$, which could be a herculean task, especially when $\X=(X_1,\ldots,X_p)$ and $\Sb=(S_1,\ldots,S_q)$ are large-dimensional. Second, the theoretical side: since the triplet $(Y,\X,\Sb)$ is mixed (not all discrete or continuous random vectors) the expression \eqref{eq:def} is not even a valid representation. The necessary reformulation is given in the next theorem.
\begin{thm} \label{thm1} 
Let $Y$ be a discrete random variable taking values $1,\ldots,k$, and $(\X,\Sb)$ be a \texttt{mixed} pair of random vectors. Then the conditional mutual information can be rewritten as
\beq \label{eq:thm1}
\MI(Y,\X \mid \Sb)\,=\,\E_{\X,\Sb}\Big[ {\rm KL} \big( p_{Y\mid \X,\Sb} \parallel p_{Y\mid \Sb} \big)  \Big],
\eeq
where Kullback-Leibler (KL) divergence from $ p_{Y\mid \X=\xb,\Sb=\sb}$ to $p_{Y\mid \Sb=\sb}$ is defined as
\beq \label{eq:kl}
{\rm KL}\big(p_{Y\mid \X,\Sb} \parallel p_{Y\mid \Sb}\big)~=~\sum_y\, p_{Y\mid \X,\Sb}(y|\xb,\sb) \,\log\left(\frac{p_{Y| \X,\Sb}(y|\xb,\sb) }{p_{Y|\Sb}(y|\sb)}\right).
\eeq
\end{thm}
To prove it, first rewrite the dependence-ratio \eqref{eq:ratio} solely in terms of conditional distribution of $Y$ as follows:
\[\dfrac{\Pr(Y=y\mid \X=\xb,\Sb=\sb)}{\Pr(Y=y\mid \Sb=\sb)}~=~\dfrac{p_{Y\mid \Xb,\Sb}(y|\xb,\sb) }{p_{Y|\Sb}(y|\sb)}\]
Next,
substitute this into \eqref{eq:def} and express it as
\[ \MI(Y,\Xb \mid \Sb)~=~ \iint_{\xb,\sb} \Bigg[  \sum_y\, p_{Y\mid \X,\Sb}(y|\xb,\sb) \,\log\left(\frac{p_{Y| \X,\Sb}(y|\xb,\sb) }{p_{Y|\Sb}(y|\sb)}\right)\Bigg] \dd F_{\X,\Sb}\]
Replace the part inside the square brackets by \eqref{eq:kl} to finish the proof. \qed

\begin{rem}
CMI measures how much information is shared only between $\Xb$ and $Y$ that is not contained in $\Sb$. Theorem \ref{thm1} makes this interpretation explicit. 
\end{rem}
\vspace{-.4em}
{\bf Estimator}. Goal is to develop a practical nonparametric algorithm for estimating CMI from $n$ i.i.d samples $\{\xb_i,y_i,\sb_i\}_{i=1}^n$ that works for large($n,p,q$) settings. Theorem \ref{thm1} immediately leads to the following estimator of \eqref{eq:thm1}:
\beq \label{eq:est}
\widehat{\MI}(Y, \Xb \mid \Sb)~=~\dfrac{1}{n}\sum_{i=1}^n \log \dfrac{  \widehat{\Pr}(Y=y_i|\xb_i,\sb_i)  }{  \widehat{\Pr}(Y=y_i|\sb_i)}.\eeq
\vskip.25em

{\bf Algorithm 1}. \textit{Conditional mutual information estimation}: the proposed ML-powered nonparametric estimation method consists of three simple steps: 

~~\texttt{Step 1}. Choose a machine learning classifier (e.g., support vector machines, random forest, gradient boosted trees, 
deep neural network, etc.), and call it \texttt{ML}$_0$. 

~~\texttt{Step 2}. Train the following two models:
\beas 
\texttt{ML.train}_{y|\xb,\sb}&\leftarrow &\texttt{ML}_0\big( Y \sim [\X, \Sb] \big)\\
\texttt{ML.train}_{y|\sb}&\leftarrow & \texttt{ML}_0\big( Y \sim \Sb \big)
\eeas

~~\texttt{Step 3}. Extract the conditional probability estimates $\widehat{\Pr}(Y=y_i|\xb_i,\sb_i)$ from $\texttt{ML.train}_{y|\xb,\sb}$, and $ \widehat{\Pr}(Y=y_i|\sb_i)$ from $\texttt{ML}_0\big( Y \sim \Sb \big)$, for $i=1,\ldots, n$.

~~\texttt{Step 4}. Return $\widehat{\MI}(Y, \Xb \mid \Sb)$ by applying formula \eqref{eq:est}.

\begin{rem}

We will be using the gradient boosting machine (\texttt{gbm}) of \cite{friedman2001} in our numerical examples (obviously, one can use other methods), whose convergence behavior is well-studied in literature \citep{breiman2004,zhang2004statistical}, where it was definitively shown that under some very general conditions, the empirical risk (probability of misclassification) of the gbm classifier approaches the optimal Bayes risk. This Bayes risk consistency property surely carries over to our conditional probability estimates in \eqref{eq:est}, which justifies the good empirical performance of our method in real datasets.
\end{rem}

\begin{rem}
Taking the base of the log in \eqref{eq:est} to be $2$, we get the measure in the unit of \textit{bits}. If the log is taken to be the natural $\log_e$, then it is in nats unit. We will use $\log_2$ in all our computation.
\vspace{-.4em}
\end{rem}
The proposed style of nonparametric estimation provides some important practical benefits: 

~~$\bullet$ Flexibility: Unlike traditional conditional independence testing procedures \citep{candes2018panning,berrett2019conditional}, our approach requires neither the knowledge of the exact parametric form of high-dimensional $F_{X_1,\ldots,X_p}$ nor the knowledge of the conditional distribution of $\Xb \mid \Sb$, which are generally \textit{unknown} in practice.

~~$\bullet$ Applicability: (i) Data-type: The method can be safely used for \textit{mixed} $\X$ and $\Sb$ (any combination of discrete, continuous, or even categorical variables). (ii) Data-dimension: The method is applicable to \textit{high-dimensional} $\X=(X_1,\ldots, X_p)$ and $\Sb=(S_1,\ldots, S_q)$.

~$\bullet$ Scalability: Unlike traditional nonparametric methods (such as kernel density or $k$-nearest neighbor-based methods), our procedure is scalable for \textit{big datasets} with large($n,p,q$).

\subsection{Model-based Bootstrap}
One can even perform statistical inference for our ML-powered conditional-mutual-information statistic. In order to test $H_0: Y \ci \X \mid \Sb$, obtain bootstrap-based p-value by noting that under the null $\Pr(Y=y|\X=\xb,\Sb=\sb)$ reduces to $\Pr(Y=y|\Sb=\sb)$. 

{\bf Algorithm 2}. \textit{Model-based Bootstrap}: The inference scheme proceeds as follows: 

~~\texttt{Step 1.} Let $$\widehat{p}_{i|\sb}~=~ \widehat{\Pr}(Y_i=1|\Sb=\sb_i),~~\text{for}~ i=1,\ldots,n$$ as extracted from (already estimated) the model $\texttt{ML.train}_{y|\sb}$ (step 2 of Algorithm 1). 

~~\texttt{Step 2.} Generate the null $Y_{n \times 1}^*=(Y^*_1,\ldots,Y^*_n)$ by
$$Y_i^* ~\leftarrow~{\rm Bernoulli}(\widehat{p}_{i|\sb}),~~\text{for}~ i=1,\ldots,n$$

~~\texttt{Step 3.} Compute $\widehat{\MI}(Y^*,\X \mid \Sb)$ using the Algorithm 1.

~~\texttt{Step 4.} Repeat the process $B$  times (say, $B=500$); compute the bootstrap null distribution, and return the p-value.

\begin{rem}
A parametric version of this inference was proposed by \cite{rosenbaum1984} in the context of observational causal study. His scheme resamples $Y$ by estimating $\Pr(Y=1|\Sb)$ using a logistic regression model. The procedure was called conditional permutation test.
\end{rem}

\subsection{A Few Examples}
\vspace{-.25em}
\begin{example}  Model: $X \sim {\rm Bernoulli}(0.5)$; $S \sim {\rm Bernoulli}(0.5)$; $Y=X$ when $S=0$ and $1-X$ when $S=1$. In this case, it is easy to see that the true $\MI(Y,X|S)=1$. We simulated $n=500$ i.i.d $(x_i,y_i,s_i)$ from this model and computed our estimate using \eqref{eq:est}. We repeated the process $50$ times to access the variability of the estimate. Our estimate is:
$$\widehat{\MI}(Y,X\mid S)~=~ 0.994 \pm 0.00234 .$$
with (avg.) p-value being almost zero. We repeated the same experiment by making $Y \sim {\rm Bernoulli}(0.5)$ (i.e., now true $\MI(Y,X|S)=0$), which yields
$$\widehat{\MI}(Y,X\mid S)~=~ 0.0022\pm 0.0017 .$$
with (avg.) pvalue being $0.820$.
\end{example} 

\begin{example} \textit{Integrative Genomics}. The wide availability of multi-omics data has revolutionized the field of biology. It is a general consensus among practitioners that combining individual omics data sets (mRNA, microRNA, CNV and DNA methylation, etc.) leads to improved prediction. However, before undertaking such analysis, it is probably worthwhile to check what is the additional information we gain from a combined analysis compared to a single-platform one. To illustrate this point, we use a Breast cancer multi-omics data that is a part of The Cancer Genome Atlas (TCGA, http://cancergenome.nih.gov/). It contain the expression of three-kinds of omics data sets: miRNA, mRNA, and proteomics from three kinds of breast cancer samples ($n=150$): Basal, Her2, and LumA. $\X_1$ is $150 \times 184$ matrix of miRNA, $\X_2$ is $150 \times 200$ matrix of mRNA, and $\X_3$ is $150 \times 142$ matrix of proteomics.
\beas 
\MI(Y,\X_2 \mid \X_1)= 0.013;& \text{p-value}=0.356 \\
\MI(Y,\X_3 \mid \X_1)= 0.0186;   & \text{p-value}= 0.235\\
\MI\big(Y,\{\X_2,\X_3\} \mid \X_1\big) = 0.0192; & \text{p-value}= 0.501.
\eeas 
It shows: neither mRNA or proteonomics add any substantial information beyond what is already captured by miRNAs.
\vspace{-.35em}
\end{example} 

%


\section{Elements of Admissible Machine Learning} \label{sec:EAML}
How to design admissible machine learning algorithms with enhanced efficiency, interpretability, and equity?\footnote{However, the general premise of admissible ML
is extremely broad and flexible, and will continue to evolve with the regulatory requirements to ensure rapid development of trustworthy algorithmic methods.} A systematic pipeline for developing such admissible ML models is laid out in this section, which is grounded in the earlier information-theoretic concepts and nonparametric modeling ideas.

\subsection{COREml: Algorithmic Interpretability} \label{sec:AINTER}

\subsubsection{From Predictive Features to Core Features} \label{sec:P2C}

One of the first tasks of any predictive modeling is to identify the key drivers that are affecting the response $Y$. Here we will discuss a new information-theoretic graphical tool to quickly spot the ``core'' decision-making variables, which are going to be vital in building interpretable models. One of the advantages of this method is that it works even in the presence of correlated features, as the following example illustrates; also see Appendix \ref{sec:Iris}.
\vskip1em

\begin{figure}[ ]
\vspace{-2em}
\centering
\begin{tikzpicture}[scale=1.4]
        \node[state] at (0, 0)      (y) {$Y$};
        \node[rectangle,draw,minimum size=1.2cm] at (6, 0)  (x50) {$X_{50}$};
        \node[state] at (3,-2.4) (x1) {$X_1$};
         \node[state] at (3,2.4) (x2) {$X_2$};
\draw
(y) edge[line width=.35mm] node {} (x1)
(y) edge[line width=.35mm] node {} (x2)
(x1) edge[line width=.35mm] node {} (x50)
(x2) edge[line width=.35mm] node {} (x50);
\end{tikzpicture}\\[4em]
    \centering
  \includegraphics[width=.48\linewidth,trim=.5cm .88cm .5cm .5cm]{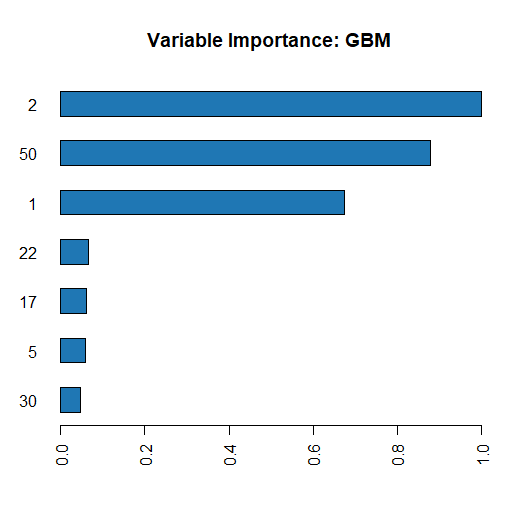}~~~~~~~
  \includegraphics[width=.478\linewidth,trim=.8cm 1.55cm 1cm .5cm]{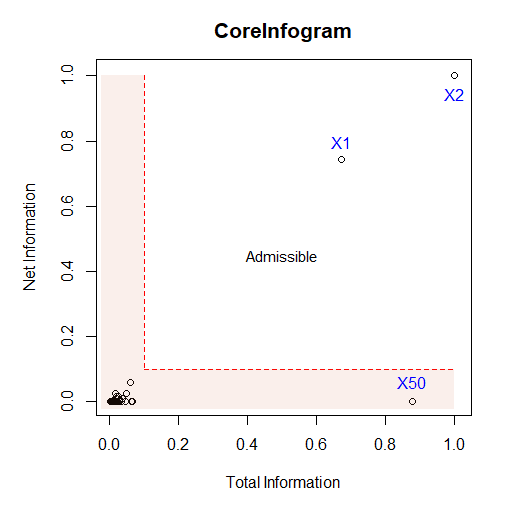}
\vskip2.75em
    \caption{Top: The graphical representation of example \ref{ex:cor} is shown. Bottom-left: The gbm-feature importance score for top seven features; rest are almost zero thus not shown. Bottom-right: infogram identifies the core variables $\{X_1,X_2\}$ from the $X_{50}$. The L-shaped area with $0.1$ width is highlighted in red; it contains inadmissible variables with either low relevance or high redundancy.} \label{fig:corex}
  \end{figure}

\begin{example} \label{ex:cor}
\textit{Correlated features}. $Y \sim {\rm Bernoulli}(\pi(\xb))$ where $\pi(\xb) = 1/(1+e^{-\mathcal{M}(\xb)})$ and
\beq \label{eq:Mcor}
\mathcal{M}(\xb)~=~3\sin(X_1) - 2X_2.\eeq
$X_1, \ldots X_{p-1}$ be i.i.d $\cN(0,1)$ random variables, and
\beq  \label{eq:xp}
X_{p}=2X_1 - X_2  + \epsilon, ~\text{where}~ \epsilon \sim \cN(0,2),\eeq
which means $X_p$ has no additional predictive value beyond what is already captured by the core variables $X_1$ and $X_2$.  Another way of saying this is that $X_p$ is \textit{redundant}---the conditional mutual information between $Y$ and $X_p$ given $\{X_1,X_2\}$ is zero:
\[
\MI\left(Y,X_p \mid \{X_1,X_2\}\right)=0.\]
The top of Fig. \ref{fig:corex} graphically depicts this. The following nomenclature will be useful for discussing our method: 
\beas 
{\rm CoreSet}&=&\{ X_1, X_2\}\\
{\rm Imitator}&=& \{ X_p\}\\
{\rm Probes}&=&\{X_3,\ldots, X_{p-1}\}.
\eeas 
\end{example} 
Note that the imitator $X_p$ is \textit{highly predictive} for $Y$ due to its association with the core variables. We have simulated $n=500$ samples with $p=50$. For each feature we compute, 
\beq \label{eq:i1} R_j~=~\text{overall relevance score of $j$th predictor},~j=1,\ldots,p.\eeq
The bottom-left corner of Fig. \ref{fig:corex} shows the relative importance scores (scaled between 0 and 1) for the top seven features using \texttt{gbm} algorithm\footnote{based on whether a particular variable was selected to split on during learning a tree, and how much it improves the Gini impurity or information gain.}, which correctly finds $\{X_1,X_2,X_{50}\}$ as the important predictors. However, it is important to recognise that this  modus operandi---irrespective of the ML algorithm---can not distinguish the `fake imitator' $X_{50}$ from the real ones $X_1$ and $X_2$. 
To enable refined characterization of the variables, we have to `add more dimension' to the classical machine learning feature importance tools.

\subsubsection{{InfoGram and L-Features}}
We introduce a tool for identification of core admissible features based on the concept of net-predictive information (NPI) of a feature $X_j$.

\begin{defn} \label{def:neti}
The net-predictive (conditional) information of $X_j$ given all the rest of the variables $\X_{-j} = \{X_1,\ldots, X_p\} \backslash \{X_j\}$ is defined in terms of conditional mutual information:
\beq \label{eq:i2}
C_j~=~\MI(Y,X_j \mid \X_{-j}), ~~\text{for}~j=1,\ldots,p.  
\eeq
For easy interpretation, we standardize $C_j$ by $\frac{C_j}{\max_j C_j}$ and convert it between 0 and 1.
Infogram, which is a abbreviation of \underline{info}rmation dia\underline{gram}, is a scatter plot of $\{( R_j,C_j)\}_{j=1}^p$ over the unit square $[0,1]^2$; see the bottom-right corner of Fig. \ref{fig:corex}. 
\end{defn}
\vspace{-.35em}
{\bf L-Features}. The highlighted L-shaped area contains features that are either irrelevant or redundant. For example, notice the position of $X_{50}$ in the plot, indicating that it is highly predictive but contains no new complementary information for the response. Clearly, there could be an opposite scenario: a variable carries valuable net individual information for $Y$, despite being moderately relevant (not ranked among the top few); see Sec. \ref{sec:Bcancer}.

\vskip.64em
\begin{rem}[Predictive Features vs. CoreSet]  Recall that in Example \ref{ex:cor}, the irrelevant feature $X_{50}$ is strongly correlated with the relevant ones $X_1$ and $X_2$ through \eqref{eq:xp}, thus violate the so-called ``irrepresentable condition''--for more details see the bibliographic notes section of \citet[p. 311]{hastie2015lassobook}. In this scenario (which may easily arise in practice), it is hard to recover the ``important'' variables using traditional variable selection methods. The bottom line is: identifying \texttt{CoreSet} is a much more difficult undertaking than merely selecting the most predictive ones. The goal of infogram is to facilitate this process of discovering the key variables that are driving the outcome.
\end{rem}

\begin{rem}[CoreML]
Two additional comments before diving into a real data examples. First, 
machine learning models based on ``core'' features (\texttt{CoreML}) show improved stability, especially when there exists considerable correlation among the features.\footnote{Numerous studies have found that many current methods like partial dependence plots, LIME, and SHAP could be highly misleading, particularly when there is strong dependence among features.} This will be demonstrated in the next two sections. Second, our approach is not tied to any particular machine learning method; it is completely model-agnostic and can be integrated with any arbitrary algorithm: choose a specific classifier $\texttt{ML}_0$ and compute \eqref{eq:i1} and \eqref{eq:i2} to generate the associated infogram. 
\end{rem}

\begin{figure}[]
\vskip1em
    \centering
  \includegraphics[width=.32\linewidth,trim=.5cm .5cm .5cm .5cm]{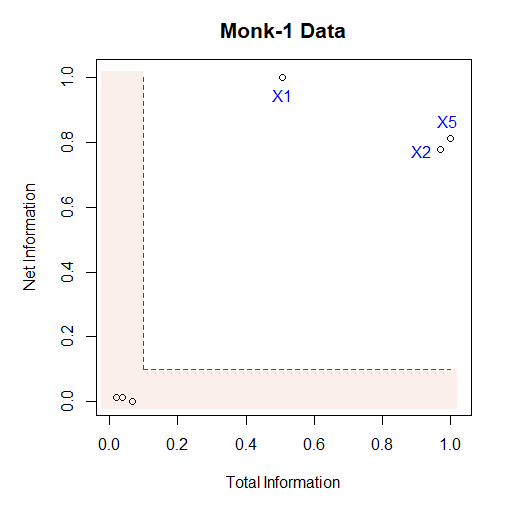}~~
    ~ \includegraphics[width=.32\linewidth,trim=.5cm .5cm .5cm .5cm]{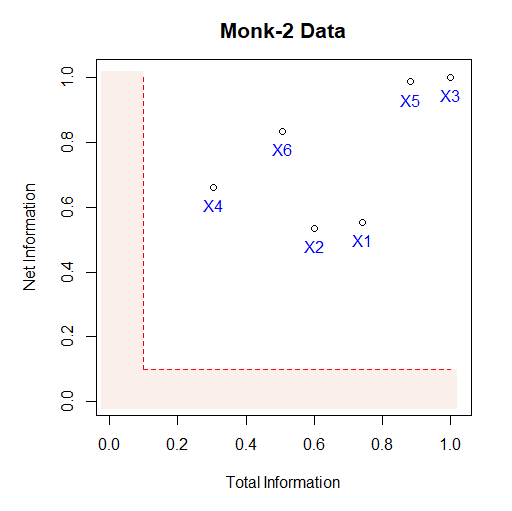}~~
  \includegraphics[width=.32\linewidth,trim=.5cm .5cm .5cm .5cm]{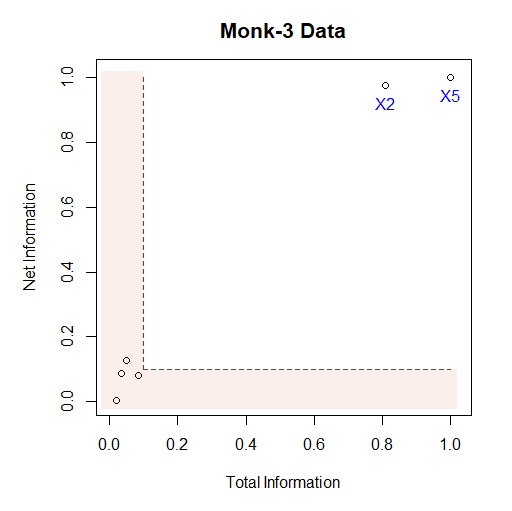}
  \vskip.5em
    \caption{Infograms of Monk's problems. CoreSets are denoted in blue.} \label{fig:monk}
  \end{figure}

\begin{example}
\textit{MONK's problems} \citep{thrun1991monk}. It is a collection of three binary artificial classification problems (MONK-1, MONK-2 and MONK-3) with $p=6$ attributes; available in the UCI Machine Learning Repository.
As shown in Fig. \ref{fig:monk}, infogram selects $\{X_1,X_2,X_5\}$ for the MONK-1 data, and $\{X_2,X_5\}$ for the MONK-3 data as the core features.
MONK-2 is an idiosyncratic case, where all six features turned out to be core! This indicates the possible complex nature of the classification rule for the MONK-2 problem.
\end{example}

\subsubsection{COREtree: High-dimensional Microarray Data Analysis} \label{sec:coretree}

How does one distill a compact (parsimonious) ML model 
by balancing accuracy, robustness, and interpretability to the best extent? To answer that, we introduce \texttt{COREtree}, whose construction is guided by infogram. The methodology is illustrated using two real datasets, namely Prostate cancer and SRBCT tumor data. The main findings are striking: it shows how one can systematically search and construct robust and interpretable shallow decision tree models (often with just two or three genes) for noisy high-dimensional microarray datasets that are as powerful as the most elaborate and complex machine learning methods.

\begin{figure}[ ]
    \centering
     \includegraphics[width=.51\linewidth,trim=.5cm .5cm .5cm 1.15cm]{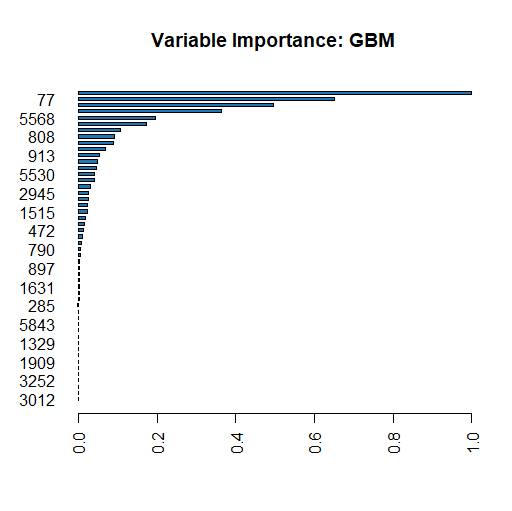}~~~~
  \includegraphics[width=.488\linewidth,trim=.5cm .5cm .5cm .5cm]{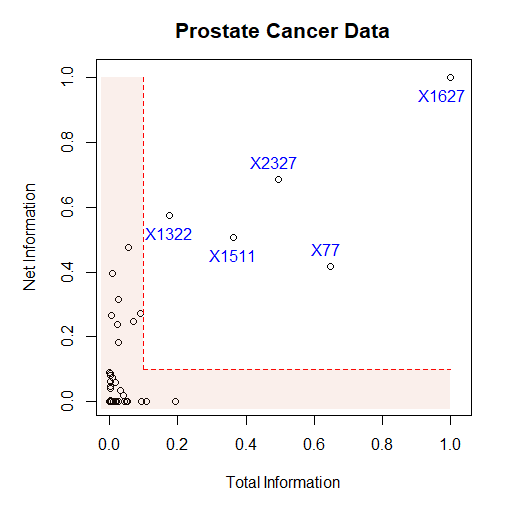}\\[2em]
  \includegraphics[width=.47\linewidth,trim=.5cm .5cm .5cm .5cm]{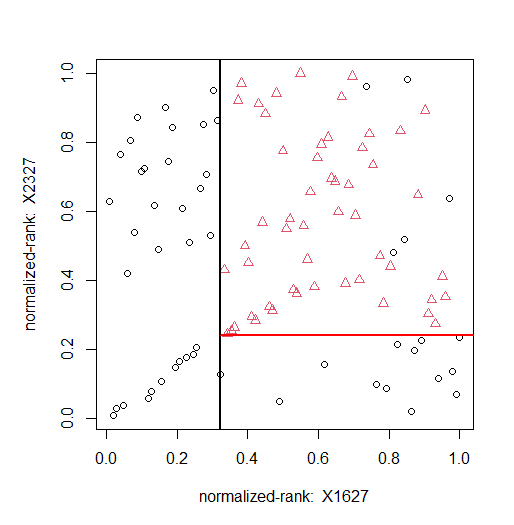}~~~~~~
  \includegraphics[width=.48\linewidth,trim=.5cm .5cm .5cm .5cm]{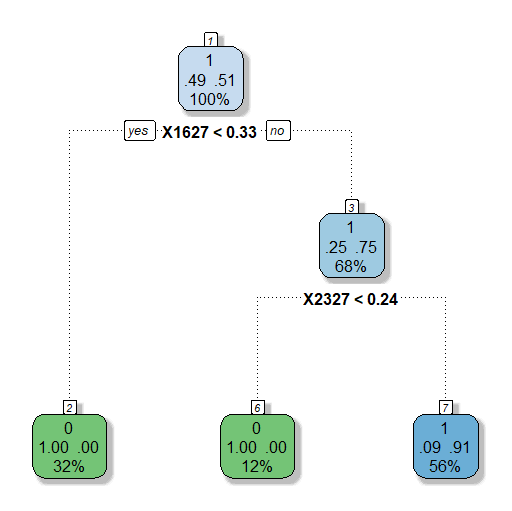}
  \vskip2em
    \caption{Prostate data analysis. Top panel: the gbm-feature importance graph, along with the infogram for the top 50 genes. Bottom-left: the scatter plot of Gene 1627 vs. 2327. For clarity, we have plotted them in the quantile domain $(u_i,v_i)$, where $u=\text{rank} (X[, 1627])/n$ and  $v=\text{rank} (X[, 2327])/n$. The black dots denote control samples with $y=0$ class and red triangles are prostate cancer samples with $y=1$ class. Bottom-right: 
    the estimated \texttt{CoreTree} with just two decision-nodes, which is good enough to be 95\% accurate.} \label{fig:Proscart}
  \end{figure}

\begin{example}
\textit{Prostate cancer gene expression data}.  The data consist of $p=6033$ gene expression measurements on $50$ control and $52$ prostate cancer patients. It is available at \texttt{https://web.stanford.edu/$\sim$hastie/CASI\_files/DATA/prostate.html}. Our analysis is summarized below.
\end{example}

~\texttt{Step} 1. \textit{Identifying CoreGenes}.  GBM-selected top 50 genes are shown in Fig. \ref{fig:Proscart}. We generate the infogram\footnote{To reduce unnecessary clutter, we have displayed the infogram using top 50 features, since the rest of the genes will be cramped inside the nonessential L-zone anyway.} of these $50$ variables (displayed on the top-right corner), which identifies five core-genes $\{1627, 2327, 77, 1511, 1322\}$. 


\vskip.4em
~\texttt{Step} 2. \textit{Rank-transform: Robustness and Interpretability}. Instead of directly operating on the gene expression values, we transform them into their ranks. Let $\{x_{j1},\ldots, x_{jn}\}$ be the measurements on $j$th gene with empirical cdf $\wtF_j$. Convert the raw $x_{ji}$ to $u_{ji}$ by
\beq u_{ji}=\wtF_j(x_{ji}),~~i=1,\ldots,n\eeq
and work on the resulting $\mathbf{U}_{n\times p}$ matrix instead of the original $\X_{n\times p}$.
We do this transformation for two reasons: first, to robustify, since it is known that gene expressions are inherently noisy. Second, to make it unit-free, since the raw gene expression values depend on the type of preprocessing, thus carries much less scientific meaning. On the other hand, percentiles are much more easily interpretable to convey ``how overexpressed a gene is.''

~\texttt{Step} 3. \textit{Shallow Robust Tree}. We build a single decision tree using the infogram-selected coregenes. This is displayed in the bottom-right panel of Fig. \ref{fig:Proscart}. Interestingly, the \texttt{CoreTree} retained only two genes $\{1627, 2327\}$ whose scatter plot (in the rank-transform domain) is shown in the bottom-left corner of Fig. \ref{fig:Proscart}. A simple eyeball estimate of the discrimination surfaces are shown in bold (black and red) lines, which closely matches with the decision tree rule.
It is quite remarkable that we have reduced the original $6033$-dimensional problem to a simple bivariate two-sample one, just by wisely selecting the features based on the infogram.

\vskip.34em
~\texttt{Step} 4.  \textit{Stability}. Note the tree that we build is based only on the infogram-selected core features. These features have less redundancy and high relevance, which provide an extraordinary stability (over different runs on the same dataset) to the decision-tree--a highly desirable characteristic.

  \vskip.34em

~\texttt{Step} 5.  \textit{Accuracy}. The accuracy of our \textit{single} decision tree (on a randomly selected 20\% test set, averaged over 100 times) is more than $95\%$. On the other hand, the full-data gbm (with $p=6033$ genes) is only $75\%$ accurate. Huge simplification of the model-architecture with significant gain in the predictive performance! 

\vskip.34em
~\texttt{Step} 6. \textit{Gene Hunting: Beyond Marginal Screening}. We compute two-sample $t$-test statistic for all $p=6033$ genes and rank them according to their absolute values (the gene with the largest absolute $t$-statistic gets ranked 1--the most differentially expressed gene). The $t$-scores for the \texttt{coregenes} along with their p-values and ranks are:
\beas \big | t_{1627} \big|=  0.15; & \text{$p$-value}=~0.88; & \text{rank}=~5383.\\   
\big | t_{2327} \big|=  1.40; & \text{$p$-value}=~0.17; & \text{rank}=~1228.
\eeas
Thus, it is hopeless to find \texttt{coregenes} by any marginal-screening method--they are \textit{too weak marginally (in isolation), but jointly an extremely strong predictor}. The good news is that our approach can find those multivariate hidden gems in a completely nonparametric fashion. 

\begin{figure}[ ]
    \centering
     \includegraphics[width=.51\linewidth,trim=1cm .6cm .5cm 1.5cm]{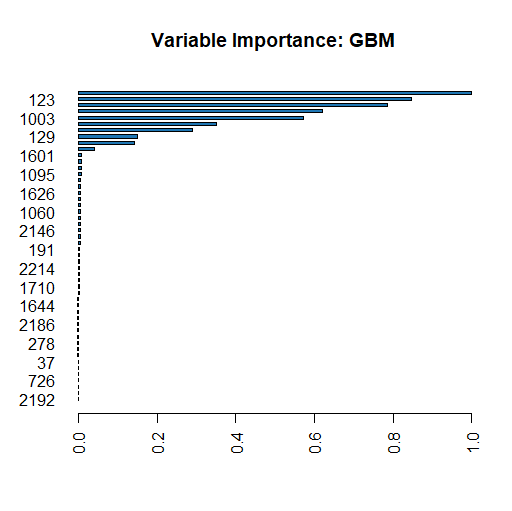}~~~~
  \includegraphics[width=.488\linewidth,trim=.5cm .5cm 1cm .5cm]{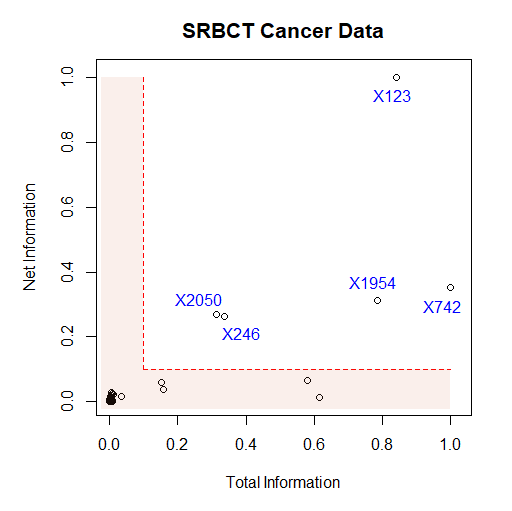}\\[3em]
  \includegraphics[width=.59\linewidth,trim=1cm .5cm 1cm .5cm]{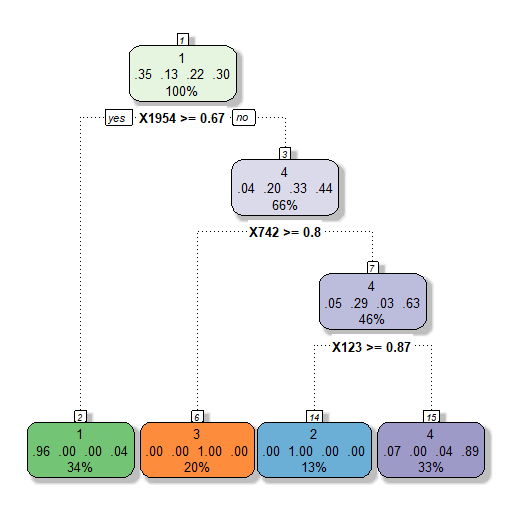}
  \vskip2em
    \caption{SRBCT data analysis. Top-left: GBM-feature importance plot; top 50 genes are shown. Top-right: The associated infogram. Bottom panel: The estimated coretree with just three decision nodes.} \label{fig:SRBCT}
  \end{figure}
  
 
\vskip.34em
~\texttt{Step} 7. \textit{Lasso Analysis and Results}. We have used the \texttt{glmnet} R-package. Lasso with $\lambda_{{\rm min}}$ (minimum cross-validation error) selects $70$ genes, where as $\lambda_{1{\rm se}}$ (the largest lambda such that error is within 1 standard error of the minimum) selects $60$ genes. Main findings are:  

\hskip2em(i) The \texttt{coregenes} $\{1627,2327\}$ were never selected, probably because they are marginally very weak; and the significant interaction is not detectable by standard-lasso. 

\hskip2em(ii) Accuracy of Lasso with $\lambda_{{\rm min}}$ is around $78\%$ (each time we have randomly selected 85\% data for training; computed the $\lambda_{{\rm cv}}$ for making prediction; averaged over 100 runs).

\vskip.4em
~\texttt{Step} 8. \textit{Explainability}. The final ``two-gene model'' is so simple and elegant that it can be easily communicated to doctors and medical practitioners: a patient with overexpressed gene $1627$ and gene $2327$ has a higher risk of getting prostate cancer. Biologists can use these two genes as robust prognostic markers for decision-making (or for recommending the proper drug). It is hard to imagine there could be a more accurate algorithm, one that is at least as compact as the ``two-gene model.'' We should not forget that the success behind this dramatic model-reduction hinges on discovering multivariate \texttt{coregenes}, which: (i) help us to gain insights into biological mechanisms [clarifying `who' and `how'], and (ii) provide a simple explanation of the predictions [justifying `why'].


\begin{example} 
\textit{SRBCT Gene Expression Data}. 
It is a microarray experiment of Small Round Blue Cell Tumors (SRBCT) taken from a childhood cancer study. It contain information on $p=2,308$ genes on $63$ training samples and $25$ test samples. Among $n=63$ tumor examples, 8 are Burkitt Lymphoma (BL), 23 are Ewing Sarcoma (EWS), 12 are neuroblastoma (NB), and 20 are rhabdomyosarcoma (RMS). The dataset is available in the \texttt{plsgenomics} R-package. The top-panel of Fig. \ref{fig:SRBCT} shows the infogram, which identifies five core genes $\{123, 742, 1954, 246, 2050\}$. The associated coretree with only three decision-nodes is shown in the bottom panel, which accurately classifies 95\% of the test cases. In addition, it enjoys all the advantages that were ascribed to the prostate data---we don't repeat them again.
\end{example}
\begin{rem}
We end this section with a general remark: when applying machine learning algorithms in scientific applications, it is of the utmost importance to design models that can clearly explain the `why and how' behind their decision-making process.  We should not forget that scientists mainly use machine learning as a \textit{tool} to gain a mechanistic understanding, so that they can  judiciously intervene and control the system. Sticking with the old way of building inscrutable predictive black-box models will severely slow down the adoption of ML methods in scientific disciplines like medicine and healthcare.
\end{rem}

\subsubsection{COREglm: Breast Cancer Wisconsin Data} \label{sec:Bcancer}
\begin{example}
\textit{Wisconsin Breast Cancer Data}. The Breast Cancer dataset is available in the UCI machine learning repository. It contains
$n=569$ malignant and benign tumor cell samples. The task is to build an admissible (interpretable and accurate) ML classifier based on $p=31$ features extracted from cell nuclei images.
\end{example}

\begin{figure}[ ]
\vskip.8em
    \centering
  \includegraphics[width=.477\linewidth,trim=1cm .5cm .5cm 1cm]{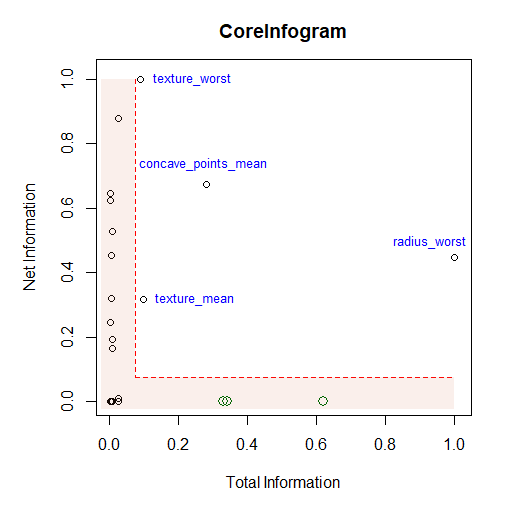}~~
    \includegraphics[width=.477\linewidth,trim=.5cm .5cm 1cm 1cm]{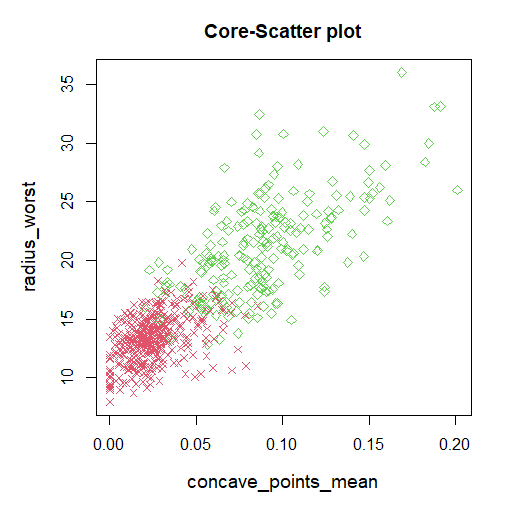}
    \vskip.65em
    \caption{Breast Cancer Wisconsin Data. Infogram reveals where the crux of the information is hidden. Infogram-guided admissible decision tree---a compact yet accurate classifier.}\label{fig:wiscon}
  \end{figure}

\texttt{Step 1.} Infogram Construction:  Fig. \ref{fig:wiscon} displays the infogram, which provides a quick understanding of the phenomena by revealing its `core.' Noteworthy points: (i) there are three highly predictive inadmissible features (green bubbles in the plot: perimeter\_worst, area\_worst, and concave\_points\_worst), which have large overall predictive importance but almost zero net individual contributions. We have called these variables `\texttt{Imitators}' in Sec. \ref{sec:P2C}. (ii) Three among the four `core' admissible features (texture\_worst, concave\_points\_mean, and texture\_mean) are not among the top features based on usual predictive information, yet they contain a considerable amount of new exclusive information (net-predictive information) that is useful for separating malignant and benign tumor cells. 
In simple terms, infogram help us to track down where the `core' discriminatory information is hidden. 

\vskip.35em
\texttt{Step 2.} Core-Scatter plot. The right panel of Fig. \ref{fig:wiscon} shows the scatter plot of the top two core features and \textit{how} they separate the malignant and benign tumor cells.

\vskip.5em
\texttt{Step 3.} Infogram-assisted \texttt{CoreGLM} model:  The simplest possible model that one could build is a logistic regression based on those four admissible features. Interestingly, the Akaike information criterion (AIC) based model selection further drops the variable \texttt{texture\_mean}, which is hardly surprising considering that it has the least net and total information among the four admissible core features. The final logistic regression model with three core variables is displayed below (output of \texttt{glm} R-function):

\vskip1.2em
\begin{lstlisting}
#COREglm Model: UCI breast cancer data
Coefficients:
                     Estimate Std. Error z value Pr(>|z|)    
(Intercept)         -29.42361    3.85131  -7.640 2.17e-14 ***
concave_points_mean  96.48880   16.11261   5.988 2.12e-09 ***
radius_worst          0.99767    0.16792   5.941 2.83e-09 ***
texture_worst         0.30451    0.05302   5.744 9.27e-09 ***
\end{lstlisting}

This simple parametric model achieves a competitive accuracy of $96.50\%$ (on a 15\% test set; averaged over $50$ trials). Compare this with full-fledged big ML models (like gbm, random forest, etc.) which attain accuracy in the range of $95-97\%$. This example again shows how infogram can guide the design of a highly transparent and interpretable \texttt{CoreGLM} model with a few handful of variables---which is as powerful as complex black-box ML methods.

\vspace{.25em}
  
\begin{rem}[Integrated statistical modeling culture]
One should bear in mind that the process by which we arrived at simple admissible models actually utilizes the power of modern machine learning---needed to estimate the formula \eqref{eq:i2} of definition \ref{def:neti}, as described by the theory laid out in section \ref{sec:theory}. For more discussion on this topic, see Appendix \ref{App:2cul} and \cite{Deep2020ISL}. In short, we have developed a process of constructing an admissible (explainable and efficient) ML procedure starting from a `pure prediction' algorithm. 
\end{rem}
\subsection{FINEml: Algorithmic Fairness} \label{sec:AFAIR}
ML-systems are increasingly used for automated decision-making in various high-stakes domains such as credit scoring, employment screening, insurance eligibility, medical diagnosis, criminal justice sentencing,  and other regulated areas. To ensure that we are making responsible decisions using such algorithms, we have to
deploy admissible models that can balance Fairness, INterpretability, and Efficiency (\texttt{FINE}) to the best possible extent. This section discusses principles and tools for designing such \texttt{FINE}-algorithms. 
\subsubsection{FINE-ML: Approaches and Limitations}
Imagine that a machine learning algorithm is used by a bank to accurately predict whether to approve or deny a loan application based on the probability of default. This ML-based risk-assessing tool has access to the following historical data:  

\hskip1.5em\sbt $Y$: \{$0, 1$\} Loan status variable--$1$ whether the loan was approved and $0$ if denied.

\hskip1.5em\sbt $\X$: Feature matrix \{income, loan amount, education, credit history, zip code\}

\hskip1.5em\sbt $\Sb$: Collection of protected attributes \{\text{gender, marital status, age, race}\}.

To automate the loan-eligibility decision-making process, the bank wants to develop an accurate classifier that will not discriminate among applicants on the basis of their protected features. Naturally, the question is: how to go about designing such ML-systems that are accurate and at the same time provide safeguards against algorithmic discrimination?

{\bf Approach 1} \textit{Non-constructive}: 
We can construct a myriad of ML models by changing and tuning different hyper-parameters, base learners, etc. One can keep building different models until one finds a perfect one that avoids adverse legal and regulatory issues.  There are at least two problems with this `try until you get it right' approach: first, it is  non-constructive. The whole process gives zero guidance on how to rectify the algorithm to make it less-biased. Second, there is no single definition of fairness---more than twenty different definitions have been proposed over the last few years \citep{narayanan21def}. And the troubling part is that these different fairness measures are mutually incompatible to each other and cannot be satisfied simultaneously \citep{kleinberg2018}; see Appendix \ref{app:sp}. Hence this laborious process could end up being a wild-goose chase, resulting in a huge waste of computation.
\vskip.5em 
{\bf Approach 2} \textit{Constructive}: Here we seek to construct ML models that---by design---mitigate bias and discrimination. To execute this task successfully, we must first identify and remove proxy variables (e.g., zip code) from the learning set, which prevent a classification algorithm from achieving desired fairness. But how to define a proper mathematical criterion to detect those surrogate variables? Can we develop some easily interpretable graphical exploratory tools to systematically uncover those problematic variables? If we succeed in doing this, then ML developers can use it as a \textit{data filtration} tool to quickly spot and remove the potential sources of biases in the pre-modeling (data-curation) stage, in order to mitigate fairness issues in the downstream analysis.

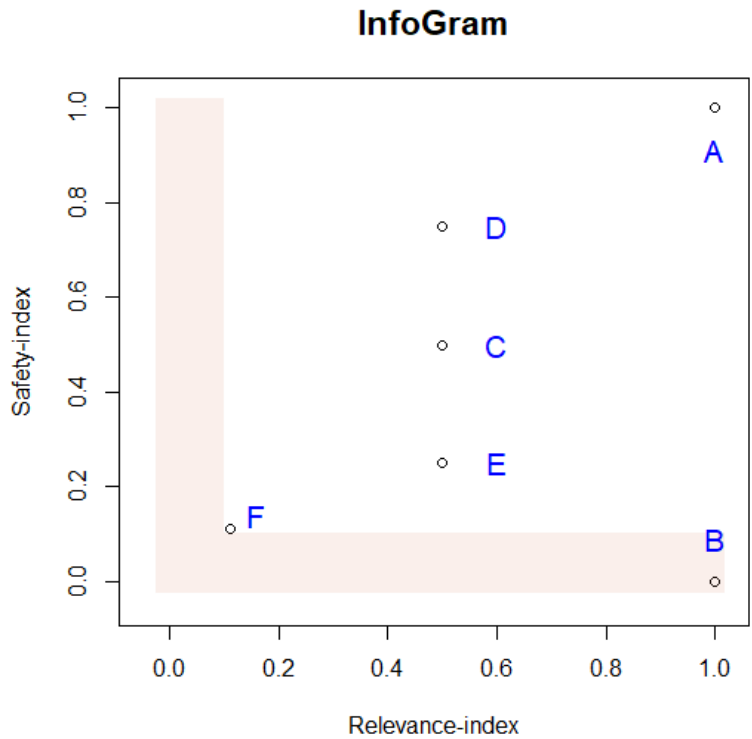
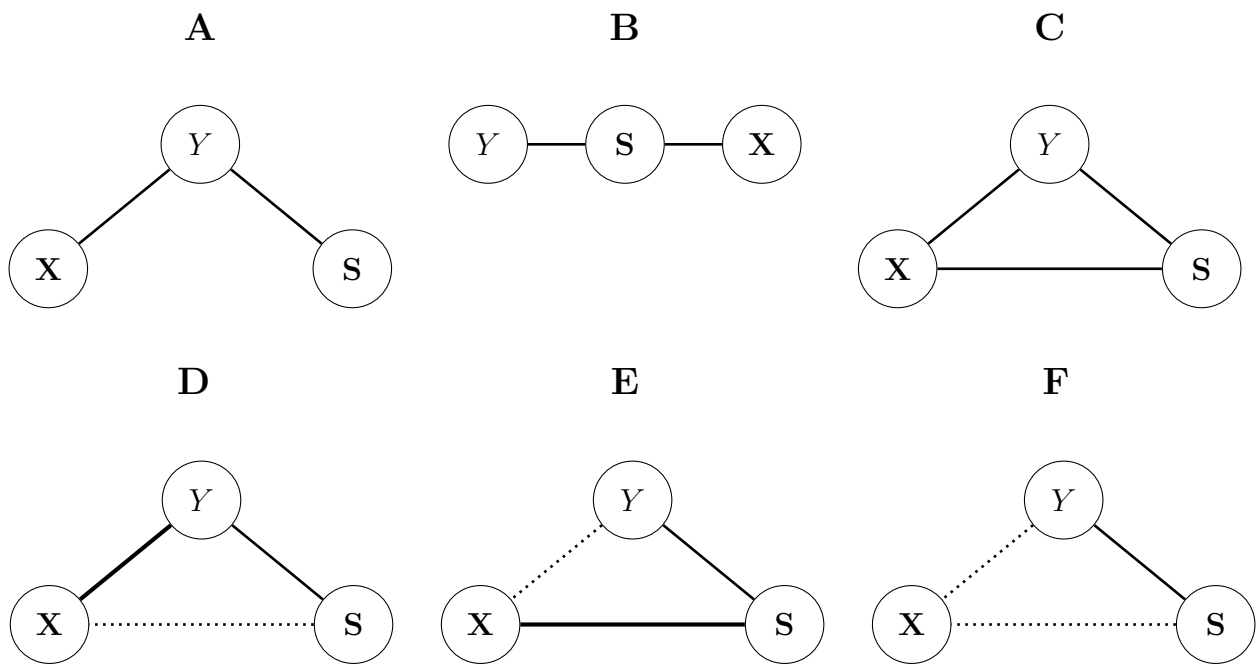
\begin{figure}
\vspace{-2em}
 \centering
  \includegraphics[width=.55\linewidth,trim=1.4cm .5cm 1.4cm 1cm]{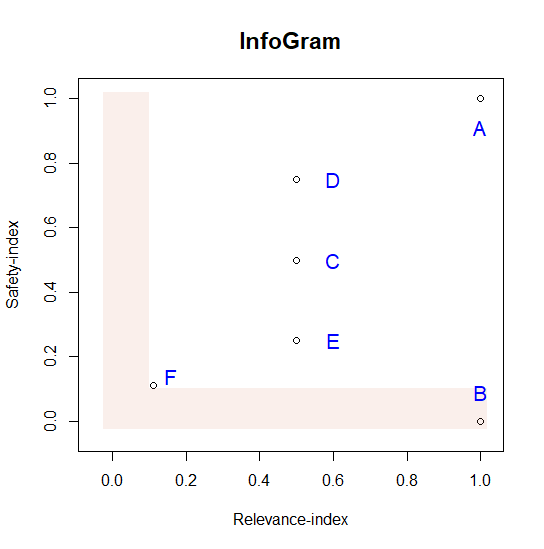}\\[.15em]
        \centering
        \begin{subfigure}[t]{.31\textwidth}
        \caption*{{\large {\bf A}} }
           \begin{center}
\begin{tikzpicture}
        \node[state] at (0, 0)      (x) {$\X$};
        \node[state] at (4, 0)     (s) {$\Sb$};
        \node[state] at (2,1.65) (y) {$Y$};
\draw
(y) edge[line width=.35mm] node {} (s)
(y) edge[line width=.35mm] node {} (x);
\end{tikzpicture}
\end{center}
        \end{subfigure}~~
        \begin{subfigure}[t]{.32\textwidth}
        \caption*{{\large {\bf B}} }
\begin{center}
\begin{tikzpicture}
\node[state] at (0, 0) (y) {$Y$};
\node[state] at (1.8,0) (s) {$\Sb$};
\node[state] at (3.6,0) (x) {$\X$};
\draw
(y) edge[line width=.35mm] node {} (s)
(s) edge[line width=.35mm] node {} (x);
\end{tikzpicture} 
\end{center}
        \end{subfigure}~~
        \begin{subfigure}[t]{.31\textwidth}
        \caption*{{\large {\bf C}} }
         \begin{center}
\begin{tikzpicture}
        \node[state] at (0, 0)      (x) {$\X$};
        \node[state] at (4, 0)     (s) {$\Sb$};
        \node[state] at (2,1.65) (y) {$Y$};
\draw
(y) edge[line width=.35mm] node {} (s)
(s) edge[line width=.35mm] node {} (x)
(y) edge[line width=.35mm] node {} (x);
\end{tikzpicture}
\end{center}   
        \end{subfigure} \\[1.6em]
     \begin{subfigure}[t]{.295\textwidth}
        \caption*{{\large {\bf D}} }
         \begin{center}
\begin{tikzpicture}
        \node[state] at (0, 0)      (x) {$\X$};
        \node[state] at (4, 0)     (s) {$\Sb$};
        \node[state] at (2,1.65) (y) {$Y$};
\draw
(y) edge[line width=.35mm] node {} (s)
(s) edge[line width=.35mm,dotted] node {} (x)
(y) edge[line width=.55mm] node {} (x);
\end{tikzpicture}
\end{center}   
        \end{subfigure}    ~~~~
 \begin{subfigure}[t]{.295\textwidth}
        \caption*{{\large {\bf E}} }
         \begin{center}
\begin{tikzpicture}
        \node[state] at (0, 0)      (x) {$\X$};
        \node[state] at (4, 0)     (s) {$\Sb$};
        \node[state] at (2,1.65) (y) {$Y$};
\draw
(y) edge[line width=.35mm] node {} (s)
(s) edge[line width=.55mm] node {} (x)
(y) edge[line width=.35mm,dotted] node {} (x);
\end{tikzpicture}
\end{center}   
        \end{subfigure}   ~~~~   
\begin{subfigure}[t]{.295\textwidth}
        \caption*{{\large {\bf F}} }
         \begin{center}
\begin{tikzpicture}
        \node[state] at (0, 0)      (x) {$\X$};
        \node[state] at (4, 0)     (s) {$\Sb$};
        \node[state] at (2,1.65) (y) {$Y$};
\draw
(y) edge[line width=.35mm] node {} (s)
(s) edge[line width=.35mm,dotted] node {} (x)
(y) edge[line width=.35mm,dotted] node {} (x);
\end{tikzpicture}
\end{center}   
        \end{subfigure}        

\vskip.88em
        \caption{Infogram maps variables in a two dimensional (effectiveness vs. safety) diagram. It is a \textit{pre-modeling} nonparametric exploratory tool for admissible feature selection. Infogram is interpreted based on graphical (conditional) independence structure. In real problems, all variables will have some degree of correlation with the protected attributes. Important part is to quantify the ``degree,'' which is measured through eq. \eqref{eq:Fval}---as indicated by varying thicknesses of the edges (bold to dotted). Ultimately, the purpose of this graphical diagnostic tool is to provide the necessary guardrails to construct an appropriate learning algorithm that can retain as much of the predictive accuracy as possible, while defending against unforeseen biases---tool for risk-benefit analysis.
       }  \label{fig:graph}
    \end{figure}

\subsubsection{InfoGram and Admissible Feature Selection}
We offer a diagnostic tool for identification of admissible features that are predictive and safe. Before going any further, it is instructive to formally define what we mean by `safe.' 

\begin{defn}[Safety-index and Inadmissibility] \label{eq:si}
Define the safety-index for variable $X_j$ as
\beq \label{eq:Fval}
F_j\,=\,\MI\big(Y,X_j \mid \{ S_1,\ldots,S_q\}\big)\eeq
This quantifies how much extra information $X_j$ carries for $Y$ that is not acquired through the sensitive variables $\Sb=(S_1,\ldots,S_q)$. For interpretation purposes, we standardize $F_j$ between zero and one by dividing by the $\max_j F_j$. Variables with ``small'' $F$-values (F-stands for fairness) will be called inadmissible, as they possess little or no informational value
beyond their use as a dummy for protected characteristics. 
\vspace{-.45em}
\end{defn}
\textit{Construction}. In the context of fairness, we construct the infogram by plotting $\{(R_j, F_j)\}_{j=1}^p$, where recall $R_j$ denotes the relevance score \eqref{eq:i1} for $X_j$. The goal of this graphical tool is to assist identification of admissible features which have little or no information-overlap with sensitive attributes $\Sb$, yet are reasonably predictive for $Y$.  
\textit{Interpretation}. Fig. \ref{fig:graph} displays an infogram with six covariates. The L-shaped highlighted region contains variables that are either inadmissible (the horizontal slice of L) or inadequate (the vertical slice of L) for prediction. The complementary set ${\rm L}^c$ comprises of the desired admissible features. 
Focus on variables \texttt{A} and \texttt{B}: both have the same predictive power, but are gained through a completely different manner. The variable \texttt{B} gathered information for $Y$ entirely through the protected features (verify it from the graphical representation of \texttt{B}), and is thus inadmissible.  On the other hand, the variable \texttt{A} carries direct informational value, having no connection with the prohibitive $\Sb$, and is thus totally admissible.
Unfortunately, though, reality is usually more complex than this clear-cut black and white \texttt{A}-\texttt{B} situation. The fact of the matter is: admissibility (or fairness, per se) is not a yes/no concept, but a matter of \textit{degree}\footnote{``Zero bias'' is an illusion. All models are biased (to a different degree), but some are admissible. The real question is how to methodically construct those admissible ones from possibly biased data.}, which is explained at the bottom two rows of Fig. \ref{fig:graph} utilizing variables C to F.

\begin{rem}
The graphical exploratory nature of the \texttt{infogram} makes the whole learning process much more transparent, interactive, and human-centered.
\vspace{-.35em}
\end{rem}

\textit{Legal doctrine}. Note that in our framework the protected variables are used only in the pre-deployment phase to determine what other (admissible) attributes to include in the algorithm to mitigate unforeseen downstream bias, which is completely legal \citep{hellman2020}.  It is also advisable that once inadmissible variables are identified using an infogram,
not to throw them (especially the highly predictive ones such as the feature B in Fig. \ref{fig:graph}) blindly from the analysis without consulting  domain experts---including some of them may not necessarily imply violation of the law; ultimately, it is up to the policymakers and judiciary to determine their appropriateness (legal permissibility) based on the given context.  Our job as statisticians is to discover those hidden inadmissible L-features (preferably in a fully data-driven and automated manner) and raise a red flag for further investigation. 


\subsubsection{FINEtree and ALFA-Test:  Financial Industry Applications} \label{sec:FINAPP}
\begin{example}
\textit{The Census Income Data}.
The dataset is extracted from 1994 United States Census Bureau database, available in UCI Machine Learning Repository. It is also known as the ``Adult Income'' dataset, which contains $n= 45,222$ records involving personal details such as yearly income (whether it exceeds \$50,000 or not), education level, age, gender, marital-status, occupation, etc. The classification task is to determine whether a person makes \$50k per year based on a set of $14$ attributes, of which four are protective:
\vskip.2em
$\hskip10em \Sb=\big\{ \texttt{Age},~\texttt{Gender},~ \texttt{Race},~ \texttt{Marital\_Status} \big\}.$
\vskip.2em
\end{example}

\texttt{Step 1}. Trust in data. Is there any evidence of built-in bias in the data? That is to say, whether a `significant' portion of the decision-making ($Y$ is greater or less than 50k per year) was influenced by the sensitive attributes $\Sb$ beyond what is already captured by other covariates $\Xb$?  One may be tempted to use $\MI(Y,\Sb \mid \Xb)$ as a measure for assessing fairness. But we need to be careful while interpreting the value of $\MI(Y,\Sb \mid \Xb)$. It can take a `small' value for two reasons:  First, a genuine case of fair decision-making where individuals with similar $\xb$ received a similar outcome irrespective of their age, gender, and other protected characteristics; see Appendix \ref{app:sp} for one such example. Second,  there is a collusion between $\Xb$ and $\Sb$ in the sense that $\Xb$ contains some proxies of $\Sb$ which \textit{reduce} its effect-size---leading one to falsely declare a decision-rule fair when it is not.
\begin{rem}[Shielding Effect]
The presence of a highly-correlated surrogate variable in the conditional set drastically reduces the size of the CMI-statistics. We call this contraction phenomenon of effect-size in the presence of proxy feature the ``shielding effect.'' To guard against this effect-distortion phenomenon we first have to identify the admissible features from the infogram.
\end{rem}

\begin{figure}[ ]
    \centering
  \includegraphics[width=.486\linewidth,trim=.5cm 1cm .5cm 1cm]{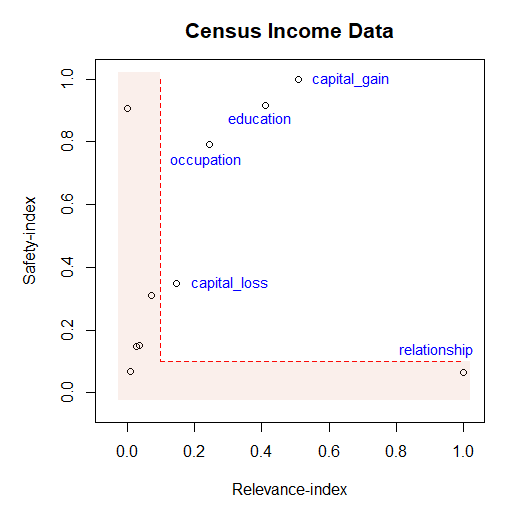}~~
  \includegraphics[width=.5\linewidth,trim=.5cm 0cm .5cm .5cm]{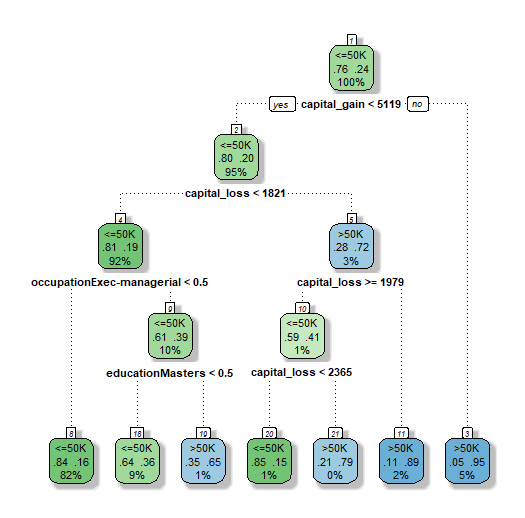}
 \vskip1em
    \caption{Census Income Data. The left plot shows the infogram. And FINEtree is displayed on the right.} \label{fig:income}
  \end{figure}

\texttt{Step 2}. Infogram to identify inadmissible proxy features. The infogram, shown in the left panel of Fig. \ref{fig:income}, finds four admissible features
\[~~~~~~\Xb_A~=~\big\{ \texttt{Capital\_gain}, ~\texttt{Capital\_loss}, ~\texttt{Occupation}, ~\texttt{Education} \big\}.~~\]
They share very little information with $\Sb$ yet are highly predictive. In other words, they enjoy
high relevance and high safety-index. Next, we also see that there is a feature that appears at the lower-right corner

$\hskip12em \Xb_R~=~\big\{ \texttt{Relationship}\big\}$
\vskip.2em
which is the prime source of bias; the subscript `R' stands for risky.  The variable \texttt{relationship} represents the respondent's role in the family---i.e., whether the breadwinner is husband, wife, child, or other relative. 
\begin{rem}
Since $\Xb_R$ is highly predictive, most \textit{unguided} ``pure prediction'' ML algorithms will include it in their models, even though it is quite unsafe. 
Admissible ML models should avoid using variables like \texttt{relationship} to reduce unwanted bias.\footnote{or at least should be assessed by experts to determine their appropriateness.}  A careful examination reveals that there could be some unintended association between \texttt{relationship} and other protected attributes due to social constructs. Without any formal method, it is a hopeless task (especially for practitioners and policymakers; see \citealt[Sec. 5.2]{lakkaraju2020fool}) to identify these innocent-looking proxy variables in a scalable and automated way. 
\vspace{-.65em}
\end{rem}

\vskip.5em
\texttt{Step 3}. ALFA-test and encoded bias. We can construct an honest fairness assessment metric by conditioning CMI with $\X_{A}$ (instead of $\Xb)$:
\beq \label{eq:ibias}
\widehat{\MI}(Y,\Sb \mid \Xb_A)=0.13,~\text{ with pvalue almost $0$}. \eeq
This strongly suggests historical bias or discrimination is encoded in the data.  Our approach not only quantifies but also allows ways to mitigate bias to create an admissible prediction rule; this will be discussed in Step 4. The preceding discussions necessitate the following, new general class of fairness metrics.
\begin{defn}[Admissible Fairness Criterion] \label{def:adf}
To check whether an algorithmic decision is fair given the sensitive attributes and the set of admissible features (identified from infogram), define
\underline{A}dmissib\underline{L}e \underline{FA}irness criterion, in short the ALFA-test, as
\beq \label{eq:alfa}
\bal_Y \,:= \,\bal(Y \mid \Sb, \Xb_A)\,=\,\MI(Y,\Sb \mid \Xb_A).~~~
\eeq 
\end{defn}
\vspace{-.4em}
{\bf Three Different Interpretations}. The ALFA-statistic \eqref{eq:alfa} can be interpreted from three different angles. 

$\bullet$ It quantifies the trade-off between fairness and model performance: how much net-predictive value is contained within $\Sb$ (and its close associates)? This is the price we pay in terms of accuracy to ensure a higher degree of fairness.

$\bullet$ A small $\bal$-inadmissibility value ensures that individuals with similar `admissible characteristics' receive a similar outcome. Note that our strategy of comparing individuals with respect to only (infogram-learned) ``admissible'' features allows us to avoid the (direct and indirect) influences of sensitive attributes on the decision making.

$\bullet$ Lastly, the $\bal$-statistic can also be interpreted as ``bias in response $Y$.''  For a given problem, if we have access to several ``comparable'' outcome variables\footnote{e.g, \cite{obermeyer2019} showed that healthcare cost can be a poor proxy of health, especially for Black patients; similarly, \cite{blattner2021costly} showed that credit scores could be a poor proxy for creditworthiness especially for low-income and minority groups.} then we choose the one which minimizes the $\bal$-inadmissibility measure. In this way, we can minimize the loss of predictive accuracy while mitigating the bias as best as we can.

\vspace{.4em}

\begin{rem}[Generalizability]
Note that, unlike traditional fairness measures, the proposed ALFA-statistic is valid for multi-class problems with a set of multivariate mixed protected attributes---which is, in itself, a significant step forward. 
\vspace{-.2em}
\end{rem}

\texttt{Step 4}. FINEtree.  The inherent historical footprints of bias (as noted in eq. \ref{eq:ibias}) need to be deconstructed to build a less-discriminatory classification model for the income data.  Fig. \ref{fig:income} shows FINEtree---a simple decision tree based on the four admissible features, which attains 83.5\% accuracy. 
\begin{rem}
FINEtree is an inherently explainable, fair, and highly competent (decent accuracy) model whose design was guided by the principles of admissible machine learning. 
\end{rem}
\vspace{-.24em}
\texttt{Step 5}. Trust in algorithm through risk assessment and ALFA-ranking: 
The current standard for evaluating ML models is primarily based on predictive accuracy on a test set, which is narrow and inadequate. For an algorithm to be deployable it has to be \textit{admissible}; an unguided ML carries the danger of inheriting bias from data.  To see that, consider the following two models:
\vskip.2em
\hskip7em Model$_{\rm A}$: ~Decision tree based on $\Xb_A$ ~(FINEtree)

\hskip7em Model$_{\rm R}$: ~Decision tree based on $\Xb_A \cup \{ \text{relationship}\}$.
\vskip.2em
Both models have comparable accuracy around $83.5\%$. Let $\whY_{\rm A}$  and $\whY_{\rm R}$ be the predicted labels based on these two models, respectively.  Our goal is to compare and rank different models based on their risk of discrimination using ALFA-statistic:
\bea 
\hbal_{\rm A} &=&\widehat{\MI}(\whY_A,\Sb \mid \Xb_A)~=~0.00042,~\text{ with pvalue $0.95$} \label{eq:al1}\\
\hbal_{\rm R} &=&\,\widehat{\MI}(\whY_{\rm R},\Sb \mid \Xb_A)~=~0.195,~\text{ with pvalue almost $0$}. \label{eq:al2}
\eea
$\bal$-inadmissibility statistic measures 
how much the final decision (prediction) was impacted 
by the protective features. A smaller value is better in the sense that it indicates improved fairness of the algorithm's decision. Eqs \eqref{eq:al1}-\eqref{eq:al2} immediately imply that Model$_{\rm A}$ is better (less discriminatory without being inefficient) than Model$_{\rm R}$, and can be safely put into production. 
\vskip.2em
\begin{rem}
Infogram and ALFA-testing can be used (by oversight board or regulators) as a fully-automated exploratory auditing tool that can systematically monitor and discover signs of bias or other potential gaps in compliance\footnote{Under the Algorithmic Accountability Act, large AI-driven corporations have to perform broader ``admissibility'' tests to keep a check on their algorithms' fairness and trustworthiness; see Appx. \ref{app:act}.}; see Appendix \ref{app:fha}.
\vspace{-1em}
\end{rem}

\vspace{.4em}
\begin{example}
\textit{Taiwanese Credit Card data}. This dataset was collected in October 2005, from a Taiwan-based bank (a cash and credit card issuer). It is available in the UCI Machine Learning Repository. We have records of $n=30,000$ cardholders, and for each we have a response variable $Y$ denoting: default payment status (Yes = 1, No = 0), along with $p=23$ predictor variables, including demographic factors, credit data, history of payment, etc. Among these $23$ features we have two protected attributes: \texttt{gender} and \texttt{age}.
\end{example}

\begin{figure}[ ]
    \centering
  \includegraphics[width=.5\linewidth,trim=.5cm .5cm .5cm -1cm]{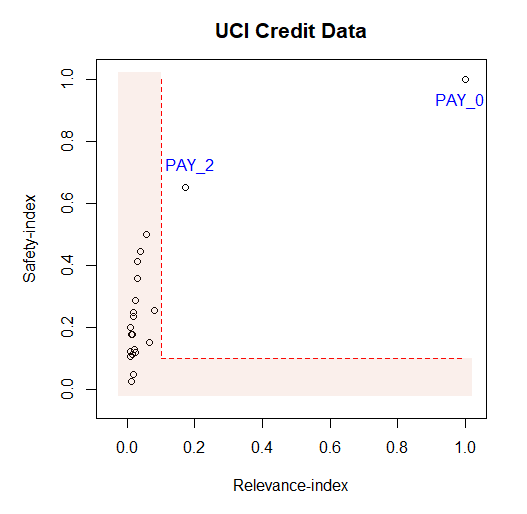}~~
  \includegraphics[width=.47\linewidth,trim=.5cm .5cm .5cm 0cm]{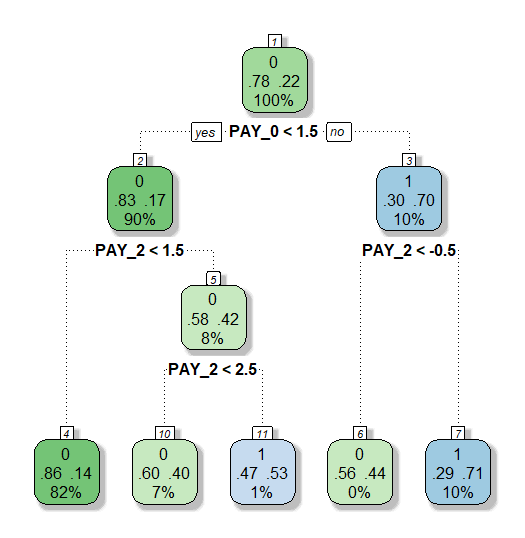}
  \vskip1em
    \caption{Left: Infogram of UCI credit card data. It selects two admissible features  (i.e., those that are relevant and less-biased) that lie in the complementary of the ``L''-shaped region. Right: The FINEtree (test data accuracy 82\%).}
\label{fig:finetree}
  \end{figure}

The infogram, shown in the left panel of Fig. \ref{fig:finetree}, clearly selects the variable \texttt{Pay\_0} and  \texttt{Pay\_2} as the key admissible factors that determine the likelihood of default. Once we know the admissible features, the next question is: `how' \texttt{Pay\_0} and \texttt{Pay\_2} are impacting the credit risk? Can we extract an admissible decision rule? For that we construct the FINEtree: a decision tree model based on the infogram-selected admissible features; see Fig. \ref{fig:finetree}.
The resulting predictive model is extremely transparent (with shallow yet accurate decision trees\footnote{One can slightly improve accuracy by combining hundreds or thousands of trees (based on only the admissible features) using random forest or boosting. But the opacity of such models renders them unfit for deployment in financial and bank sectors \citep{fahner2018fico}.}) and also mitigates unwanted bias by avoiding inadmissible variables. Lenders, regulators, and bank managers can use this model for automating credit decisions.

\begin{figure}[ ]
    \centering
  \includegraphics[width=.5\linewidth,trim=.5cm 1cm .5cm 0cm]{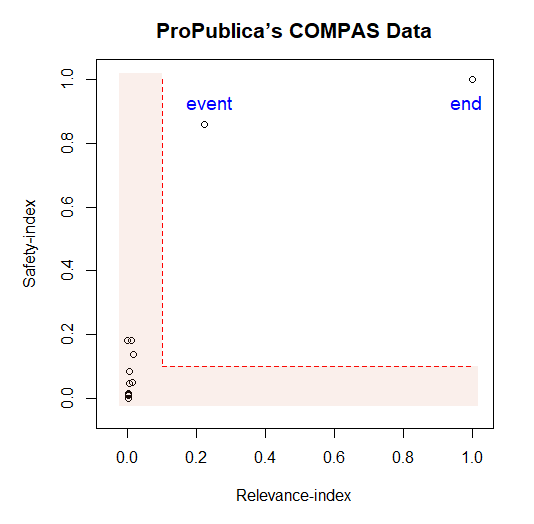}~~~
  \includegraphics[width=.46\linewidth,trim=.5cm 0cm .5cm 1.5cm]{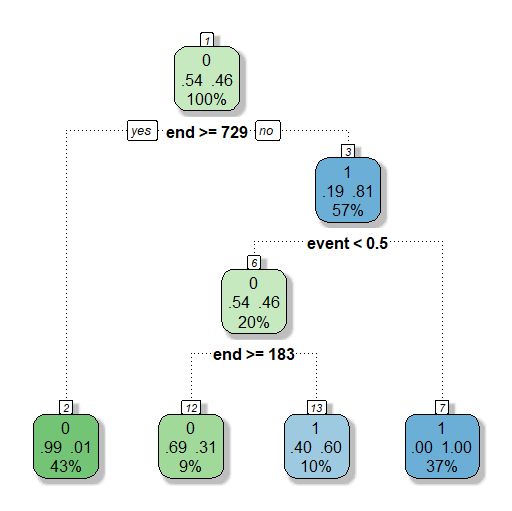}\\[5em]
  \includegraphics[width=.488\linewidth,trim=.5cm 1cm .5cm .5cm]{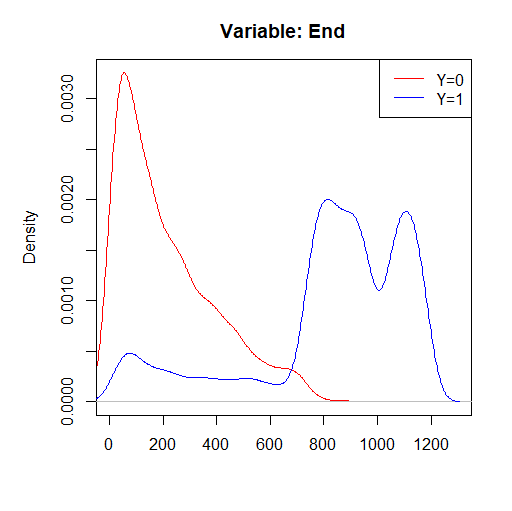}~~~\includegraphics[width=.48\linewidth,trim=.5cm .75cm .5cm 1.25cm]{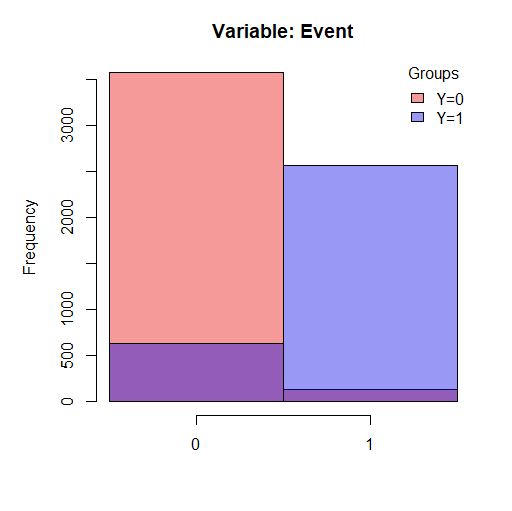}
  \vskip1.4em
    \caption{ProPublica’s COMPAS Data: Top row: infogram and the estimated FINEtree. Bottom row: The two-sample distribution of the continuous variable \texttt{end} and binary \texttt{event} show their usefulness for predicting whether a defendant will recidivate or not.} \label{fig:compas}
  \end{figure}

\subsubsection{Admissible Criminal Justice Risk Assessment}
\begin{example}
\textit{ProPublica’s COMPAS Data}. COMPAS--an acronym for Correctional Offender Management Profiling for Alternative Sanctions--is a most widely used commercial algorithm within the criminal justice system for predicting recidivism risk (the likelihood of re-offending). The data\footnote{Data: https://github.com/propublica/compas-analysis/raw/master/compas-scores-two-years.csv}---complied by a team of journalists from ProPublica---constitute all criminal defendants who were subject to COMPAS screening  in Broward County, Florida, during 2013 and 2014. For each defendant, $p=14$ features were gathered, including demographic information, criminal history, and other administrative information. Besides, the dataset also contains information on whether the defendant did in fact actually recidivate (or not) within two years of the COMPAS administration date (i.e., through the end of March 2016); and 3 additional sensitive attributes (gender, race, and age) for each case.

\vskip.25em
The goal is to develop a accurate and fairer algorithm to predict whether a defendant will engage in violent crime or fail to appear in court if released. Fig. \ref{fig:compas} shows our results. Infogram selects \texttt{event} and \texttt{end} as the vital admissible features. The bottom row of Fig. \ref{fig:compas} confirms their predictive power. Unfortunately, these two variables are not explicitly defined by ProPublica in the data repository. Based on \cite{brennan2009evaluating}, we feel that \texttt{event} indicates some kind of crime that resulted in a prison sentence during a past observation period
(we suspect the assessments were conducted by local probation officers during some period between January 2001 and
December 2004), and the variable \texttt{end} denotes the number of days under observation (first event or end of study, whichever occurred first). The associated FINEtree recidivism algorithm based on \texttt{event} and \texttt{end} reaches 93\% accuracy with AUC $0.92$ on a test set (consist of 20\% of the data). Also see Appendix \ref{sec:compas2}.
\vspace{-.3em}
\end{example}

\subsubsection{FINEglm and Application to Marketing Campaign}
We are interested in the following question: how does one systematically build fairness-enhancing parametric statistical algorithms, such as a generalized linear model (GLM)?

\begin{example}
\textit{Thera Bank Financial Marketing Campaign}. This is a case study about Thera Bank, the majority of whose customers are liability customers (depositors) with varying sizes of deposits---and among them, very few are borrowers (asset customers). The bank wants to expand its client network to bring more loan business and in the process, earn more through the interest on loans. To test the viability of this business idea they ran a small marketing campaign with $n=5000$ customers where a $480$ (= 9.6\%) accepted the personal loan offer. Motivated by the healthy conversion rate, the marketing department wants to devise a much more targeted digital campaign to boost loan applications with a minimal budget.
\vspace{.5em}

\textit{Data and the problem}. For each of $5000$ customers, we have binary response $Y$: customer response to the last personal loan campaign, and $12$ other features like  customer's annual income, family size, education level, value of house mortgage if any, etc.  Among these $12$ variables, there are two protected features: \texttt{age} and \texttt{zip code}. We consider zip code as a sensitive attribute, since it often acts as a proxy for race. 

Based on this data, we want to devise an automatic and \textit{fair} digital marketing campaign that will maximize the targeting effectiveness of the advertising campaign while minimizing the discriminatory impact on protected classes to avoid legal landmines.
\end{example}

\textit{Customer targeting using admissible machine learning}. Our approach is summarized below:

\begin{figure}[ ]
    \centering
  \includegraphics[width=.49\linewidth,trim=1cm 1cm .35cm .75cm]{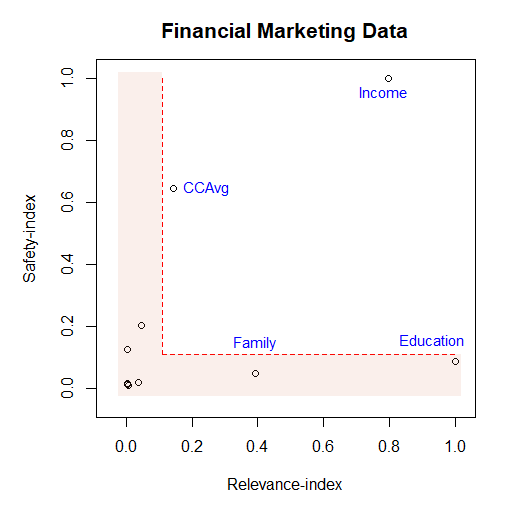}~
  \includegraphics[width=.492\linewidth,trim=.35cm 1cm 1cm .75cm]{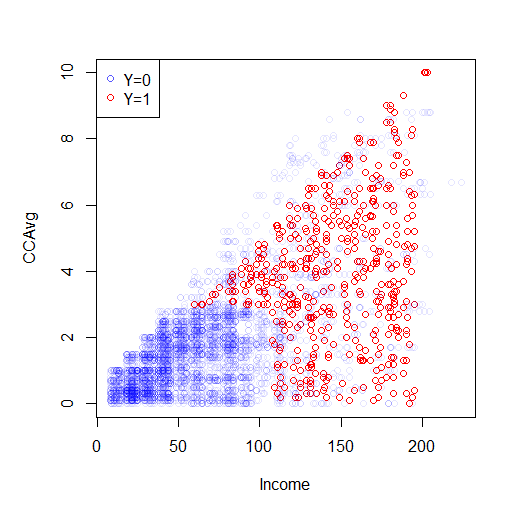}
  \vskip1.24em
    \caption{Thera Bank marketing campaign data. Left: infogram. Right: scatter plot based on the two admissible features; the color blue and red indicate two different classes.} \label{fig:marketing}
  \end{figure}

\texttt{Step 1}. Graphical tool for algorithmic risk management. Fig. \ref{fig:marketing} shows the infogram, which identifies two admissible features for loan decision: \texttt{Income} (annual income in \$000), and 
\texttt{CCAvg} (Avg. spending on credit cards per month). However, the two highly predictive variables \texttt{education} (education level: undergraduate, graduate, or advanced) and \texttt{family} (family size of the customer) turn out to be inadmissible, even though they look completely ``reasonable'' on the surface. Consequently, including these variables in a model can do more harm than good by discriminating against minority applicants. 

\begin{rem}
It is evident that infogram can be used as an algorithmic risk management tool to quickly identify and combat unwanted hidden bias. Financial regulators can use infogram to quickly spot and remediate issues of historic discrimination; see Appendix \ref{app:fha}.
\end{rem}

\begin{rem}
Infogram runs a `combing operation' to 
distill down a large, complex problem to its \textit{core} that holds the bulk of the ``admissible information.'' In our problem, the useful information is mostly concentrated into two variables---Income and CCAvg, as seen in the scatter diagram.
\end{rem}

\texttt{Step 2}. FINE-Logistic model: We train a logistic regression model based on the two admissible features, leading to the following model:
\beq 
\logit \left\{\mu(x)\right\} \,=\,-6.13\,+\,.04 \,\texttt{Income}\,+\,.06 \,\texttt{CCAvg},
\eeq
where $\mu(x)=\Pr(Y=1|\Xb=\xb)$. This simple model achieves 91\% accuracy. It provides a clear understanding of the `core' factors that are driving the model's recommendation. 

\begin{rem}[Trustworthy algorithmic decision-making]
FINEml models provide a transparent and self-explainable algorithmic decision-making system that comes with protection against unfair discrimination---which is essential for earning the trust and confidence of customers. The financial services industry can immensely benefit from this tool.
\end{rem}

\texttt{Step 3}. \texttt{FINElasso}. One natural question would be, How can we extend this idea to high-dimensional glm models? In particular, we are interested in the following question: Is there any way we can directly embed `admissibility' into the lasso regression model? The key idea is as follows:  use adaptive regularization by choosing the weights to be the inverse of safety-indices, as computed in formula \eqref{eq:Fval} of definition \ref{eq:si}. Estimate \texttt{FINElasso} model by solving the following adaptive version:
\beq \label{eq:flasso}
\widehat{\bb}_{{\rm FINE}}\,=\, \argm_{\bb} \sum_{i=1}^n \Big[ -y_i(\xb_i^T\bb)+ \log\big(1+e^{\xb_i^T\bb}\big)  \Big] ~-~ \lambda \sum_{j=1}^p w_j \big| \beta_j \big|,
\eeq 
where the weights are defined as
\beq  \label{eq:weight}
w_j^{-1}\,=\MI\big(Y,X_j \mid \{ S_1,\ldots,S_q\}\big).\eeq
The adaptive penalization in \eqref{eq:flasso} acts as a bias-mitigation mechanism by dropping (that is, heavily penalizing) the variables with very low safety-indices. This whole procedure can be easily implemented using the \texttt{penalty.factor} argument of \texttt{glmnet} R-package \citep{friedman2010regularization}. No doubt a similar strategy can be adopted for other regularized methods such as ridge or elastic-net. For an excellent review on different kinds of regularization procedures, see \cite{hastie2020ridge}.

\begin{rem}
A full lasso on $\Xb$ selects the strong surrogates (variables \texttt{family} and \texttt{education}) as some of the top features due to their high predictive power, and hence carries enhanced risk of being discriminatory.
On the other hand, an infogram-guided \texttt{FINElasso} provides an automatic defense mechanism for combating bias without significantly compromising accuracy. 
\end{rem}

\begin{rem}[Towards A Systematic Recipe]
This idea of data-adaptive `re-weighting' as a bias mitigation strategy, can be easily translated to other types of machine learning models. For example, to incorporate fairness into the traditional random forest method, choose splitting variables at each node by performing weighted random sampling.
The selection probability is determined by 
\beq \Pr(\text{selecting variable $X_j$})\,=\,\frac{F_j}{\sum_j F_j},\eeq
where the F-values $F_j$ is defined in equation \eqref{eq:Fval}. This can be easily operationalized using the \texttt{mtry.select.prob} argument of the randomForest() function in \texttt{iRF} R-package. Following this line of thought, one can (re)design a variety of less-discriminatory ML techniques without changing a single architecture of the original algorithms.
\vspace{-.3em}
\end{rem}

\section{Conclusion} \label{sec:con}
\begin{quote}
   \textit{Faced with the profound changes that AI technologies can produce, pressure for
“more” and “tougher” regulation is probably inevitable.} ~~\citep{stone2019one}.
\end{quote} %

Over the last $60$ years or so---since the early 1960s---there's been an explosion of powerful ML algorithms with increasing predictive performance. However, the challenge for the next few decades will be to develop sound theoretical principles and computational mechanisms that \textit{transform} those conventional ML methods into more safe, reliable, and trustworthy ones.

The fact of the matter is that doing machine learning in a `responsible' way is much harder than developing another complex ML technique. A highly accurate algorithm that does not comply with regulations is (or will soon be) unfit for deployment, especially in safety-critical areas that directly affect human lives. For example, the Algorithmic Accountability Act\footnote{Also see, EU's ``Artificial Intelligence Act” released on April 21, 2021, whose key points are summarized in Appendix \ref{sec:EUAI}.} (see Appx. \ref{app:act}) introduced in April 2019 requires large corporations (including tech companies, as well as banks, insurance, retailers, and many other consumer businesses) to be cognizant of the potential for biased decision-making due to algorithmic methods; otherwise, civil lawsuits can be filed against those firms. As a result, it is becoming necessary to develop tools and methods that can provide ways to \textit{enhance} interpretability and efficiency of classical ML models while guarding against bias. With this goal in mind, this paper introduces a new kind of statistical learning technology and information-theoretic automated monitoring tools that can guide a modeler to \textit{quickly} build ``better'' algorithms that are less-biased, more-interpretable, and sufficiently accurate.

\vskip.3em
One thing is clear: rather than being passive recipients of complex automated ML technologies, we need more general-purpose statistical \textit{risk management tools} for algorithmic accountability and oversight. This is critical to the responsible adoption of regulatory-compliant AI-systems. This paper has taken some important steps towards this goal by introducing the concepts and principles of `Admissible Machine Learning.'

\vskip.5em
\section*{Acknowledgement} 
The author thanks the editor, associate editor, and four anonymous reviewers for their helpful suggestions. I would like to specially thank Erin LeDell for bringing this problem to my attention. The author was benefited from many useful discussions with Michael Guerzhoy, Hany Farid, Julia Dressel, Beau Coker, and Hanchen Wang on demystifying some aspects of COMPASS data; Daniel Osei on the data pre-processing steps of Lending Club loan data. This research was supported by \texttt{H2O.ai}.

\bibliographystyle{Chicago}
\bibliography{ref-bib}

\clearpage
\newpage 
\vskip1em

\section{Appendix}
\renewcommand{\theequation}{6.\arabic{equation}}
\appendix
\renewcommand{\thesubsection}{A.\arabic{subsection}}
\subsection{Proof of Theorem 1} \label{app:proof1}
The conditional entropy $H(Y\mid \Xb, \Sb)$ can be expressed as
\bea
H(Y\mid \Xb, \Sb) &=& \iint_{\xb,\sb} H(Y \mid \Xb=\xb, \Sb=\sb) \dd F_{\xb,\sb}\nonumber\\
&=&\iint_{\xb,\sb} \Bigg\{- \int_y f_{Y|\X, \Sb}(y,\xb|\sb) \log \left( f_{Y|\X,\Sb}(y,\xb|\sb) \right) \dd y\Bigg\} \dd F_{\xb,\sb}\nonumber\\
&=&-\iiint_{\xb,\sb,y} \log \left( f_{Y|\X,\Sb}(y,\xb|\sb) \right) \dd F_{\xb,\sb,y}. \label{eq:p1}
\eea
Similarly, 
\bea
H(Y\mid \Sb) &=& \int_{\sb} H(Y \mid \Sb=\sb ) \dd F_{\sb}\hskip14em \nonumber\\
&=&  \int_{\sb} \Big\{- \int_y f_{Y|\Sb}(y|\sb) \log \left( f_{Y|\Sb}(y|\sb) \right) \dd y\Big\} \dd F_{\sb}\nonumber\\
&=&- \iiint_{\xb,\sb,y} \log \left( f_{Y|\Sb}(y|\sb) \right)   \dd F_{\xb,\sb,y}. \label{eq:p2}
\eea
Take the difference $H(Y|\Sb) - H(Y|\Xb, \Sb)$ by substituting \eqref{eq:p2} and \eqref{eq:p1} to complete the proof.   \qed
\subsection{The Algorithmic Accountability Act} \label{app:act}
This bill\footnote{https://www.congress.gov/bill/116th-congress/house-bill/2231/all-info} was introduced by Senators Cory Booker (D-NJ) and Ron Wyden (D-OR) in the Senate and Rep. Yvette Clarke (D-N.Y.) in the House on April, 2019.  It requires large companies to conduct automated decision system impact assessments of their algorithms. Entities that develop, acquire, and/or utilize AI must be cognizant of the potential for biased decision-making and outcomes resulting from its use, otherwise civil lawsuits can be filed against those firms.  Interestingly, on Jan. 13, 2020, the Office of Management and Budget released a draft memorandum\footnote{The draft memo is available at:{\small whitehouse.gov/wp-content/uploads/2020/01/Draft-OMB-Memo-on-Regulation-of-AI-1-7-19.pdf}} to make sure the federal government doesn't over-regulate industry's AI to the extent that it hampers innovation and development.

\subsection{Fair Housing Act’s Disparate Impact Standard} \label{app:fha}
Detecting inadmissible (proxy) variables can be used as a first defense against algorithmic bias. Consider the Fair Housing Act’s Disparate Impact Standard\footnote{https://www.govinfo.gov/content/pkg/FR-2019-08-19/pdf/2019-17542.pdf} (U.S. Aug. 19, 2019)--according to §100.500 (c)(2)(i) of the Act, a defendant can rebut a claim of discrimination by showing that “none of the factors used in the algorithm rely
in any material part on factors which are substitutes or close proxies for protected classes under the Fair Housing Act.” Therefore regulators, judges, and model developers can use \texttt{infogram} as a statistical diagnostic tool to keep a check on 
the algorithmic disparity of automated decision systems. 
\subsection{Beware of The ``Spurious Bias'' Problem}
\label{app:sp}
Using a real data example, here we alert practitioners some of the flaws of current fairness criteria and discuss their remedies. Consider the admission data shown in Table \ref{tab:admission}. We are interested to know: is there a gender bias in the admission process?

\vskip.4em
$\bullet$ Marginal analysis:   the overall acceptance rate in two departments for female applicants is 37\%, whereas for male applicants it is roughly 50\%. The disparity can be quantified using the adverse impact ratio (AIR), also known as disparate impact:
\beq \label{eq:DM}
{\rm AIR}(Y,G)~=~ \dfrac{\Pr(Y=1\mid G={\rm female})}{\Pr(Y=1\mid G={\rm male})}~=~\dfrac{.37}{.50}=0.74 < 0.80\eeq
The conventional ``80\% rule''\footnote{The US Equal Employment Opportunity Commission states that fair employment should abide the 80\% rule: the acceptance rate for any group should be no less than 80\% of that of the highest-accepted group.} indicates that the admission process is biased.

\begin{table}[t]
    \centering
    \begin{tabularx}{.9\linewidth}{qqYY}
    \toprule
         Dept\newline(D)~~~~ & Gender\newline(G)~~~~ & ~Admitted\newline($y=1$) & ~~Rejected\newline($y=0$)\\
         \midrule
         \multirow{2}{*}{I} & Male& 353 &207\\
         & Female &17 & 8\\
         \midrule
         \multirow{2}{*}{II}  &Male& 138 & 279 \\
         & Female & 131 &244\\
         \bottomrule
    \end{tabularx}
    \vskip.5em
    \caption{Admission data classified by gender and departments. This is actually a part of the 1973 UC Berkeley graduate admission data; here, for simplicity, we have taken the data of Departments B and D.}    \label{tab:admission}
        \vskip.5em
\end{table}

\vskip.4em
$\bullet$ The bias-reversal phenomena:
admission chances within Department I: Male 63\% (male), and female 68\%; within Department II: Male 33\%, and female F 35\%. Thus, when we investigate the admissions by department, the discrimination against women vanishes; in fact, the bias gets reversed (in the favor of women)!
\vskip.4em
$\bullet$ Department-specific ``subgroup'' analysis:  Here we investigate the adverse impact ratio (AIR) within each department. 

For Dept I (no bias): 
\beq  \label{eq:air1} 
{\rm AIR}(Y,G\mid D={\rm I})~=~ \dfrac{\Pr({\rm Y=1}\mid G={\rm male})}{\Pr({\rm Y=1}\mid G={\rm female})}~=~.63/.68=0.92\,>\,0.80.\eeq
For Dept II (no bias):
\beq \label{eq:air2}
{\rm AIR}(Y,G\mid D={\rm II})~=~ \dfrac{\Pr({\rm Y=1}\mid G={\rm male})}{\Pr({\rm Y=1}\mid G={\rm female})}~=~.33/.35=0.94\,>\,0.80.\eeq
Eqs. \eqref{eq:DM}-\eqref{eq:air2} present us with a paradoxical situation. What will be our final conclusion on the fairness of the admission process? How to resolve it in a principled way?

\vskip.4em
$\bullet$ A resolution: Compute a measure of overall (university-wide) discrimination by ALFA-statistic (see definition \ref{def:adf} for more details):
\beq \label{eq:ami}
\bal_Y \,:= \,\MI(Y,G|D)~=~\sum_{d=0}^1\Pr(D=d) \MI(Y,G|D=d),\eeq
where $\bal$-inadmissibility statistic measures the discrimination (how predictive the admission variable $Y$ is based on gender $G$) in a particular department's admission. Applying the formula \eqref{eq:thm1} we get
\[\widehat{\bal}_Y =\widehat{\MI}(Y,G|D)=0.000285,~\,\text{with p-value:~$0.715$}.\]
This suggests $Y \ci G \mid D$, i.e., 
the gender contains no additional predictive information for admission beyond what is already captured by the department variable. The apparent gender bias can be `explained away' by the choice of the department. Graphically, this can be represented as a Markov chain:
\begin{center}
\vskip.45em
\begin{tikzpicture}[scale=1.1]
\node[state] at (0, 0) (y) {Y};
\node[state] at (2.2,0) (s) {D};
\node[state] at (4.4,0) (x) {G};
\draw
(y) edge[line width=.35mm] node {} (s)
(s) edge[line width=.35mm] node {} (x);
\end{tikzpicture} 
\end{center}
Note that there is no direct link between the gender (G) and admission (Y). Conclusion: there is no evidence of any direct sex-discrimination in the admission process.

\vskip.24em

$\bullet$ Improved AIR measure:  
one can generalize the (marginal) adverse impact ratio \eqref{eq:DM} to the following conditional one (which is similar in spirit to eq. \eqref{eq:ami}):
\beq \label{eq:cDM}
{\rm CAIR}(Y,G|D)~=~\int {\rm AIR}(Y,G|D=d) \dd F_D,\eeq 
which, in this case, can be decomposed as 
\beq \label{eq:cDMe}
{\rm CAIR}(Y,G|D)~=~\Pr(D={\rm I}) {\rm AIR}(Y,G|D={\rm I}) ~+~\Pr(D={\rm II}) {\rm AIR}(Y,G|D={\rm II}).
\eeq
Applying \eqref{eq:cDMe} for our Berkeley example data yields the following estimate:
\beas \widehat{{\rm CAIR}}(Y,G|D)&=&0.43\times 0.92 + 0.57\times 0.94\\
&=&0.93 ~>~0.80.\eeas
This shows no evidence of sex bias in graduate admissions! The moral is: beware of spurious bias, and be aware of two types of errors that might occur due to an incorrect fairness-metric: falsely rejecting a fair algorithm as unfair (Type-I fairness error), and falsely accepting an unfair algorithm as fair (Type-II fairness error).

\subsection{Revisiting COMPAS Data}
\label{sec:compas2}
\vspace{-.5em}
There is another version of the COMPAS data\footnote{{\small https://raw.githubusercontent.com/Jimmy-Lin/TreeBenchmark/master/datasets/compas/data.csv}} (binarized features) that researchers have used for evaluating the accuracy of their algorithms. This dataset contains a list of hand-picked $p=22$ features over $n=10,747$ criminal records. Goal is to build an interpretable and accurate recidivism prediction model. Infogram-selected COREtree is displayed below.

\begin{figure}[ ]
\vspace{-.5em}
    \centering
  \includegraphics[width=.52\linewidth,trim=1cm .5cm 1cm .5cm]{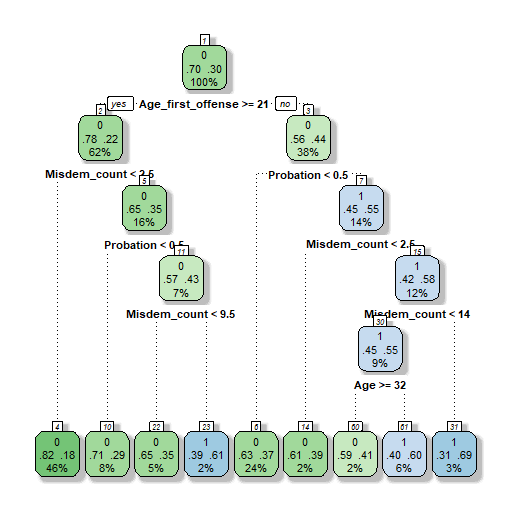}
    \caption{Infogram-selected COREtree.}
    \vspace{-.5em}\label{fig:compass}
  \end{figure} 
10-fold cross-validation shows $(72 \pm 1.50)\%$ classification accuracy of our model, which is close to the best known performance on this version of the COMPAS data.
\vspace{-.65em}
\subsection{Two Cultures of Machine Learning} \label{App:2cul}
\vspace{-.35em}
Black-box ML culture: it builds large complex models, keeping solely the predictive accuracy in mind. White-box ML culture: it directly builds interpretable models, often by enforcing domain-knowledge-based constraints on traditional ML algorithms like decision tree or neural net. Orthodox `black-or-white thinkers' of each camp have been at loggerheads for some time. This raises the question: is there any way to get the best of both worlds? If so, how? 

\textit{An Integrated (third?) culture}: In this paper, we have taken the middle path between two extremes. We leverage (instead of boycotting) the power (scalability and flexibility) of modern machine learning methods by viewing them as a heavy-duty ``toolkit'' that can efficiently drill through big complex datasets to systematically search for the hidden admissible models.

\subsection{COREtree: Iris Data} \label{sec:Iris}
The dataset includes 
three kinds of iris flowers (setosa, versicolor, or virginica) with 50 samples from each class. The task is to develop a model (preferably a compact model based on only important features) to accurately classify iris flowers based on their sepals and petals' length and width ($p=4$). Before we start our analysis, it is important to be aware of the highly-correlated nature of the 4-features; the estimated $4\times 4$ correlation matrix is displayed below: 
\beq \label{iris:cor}
\vspace{-.05em}
{\hat\Sigma}_\rho~=~\begin{bmatrix}
1.000      &-0.118        &{\bf 0.872}      &{\bf 0.818}\\
-0.118       &1.000       &-0.428      &-0.366\\
{\bf 0.872}      &-0.428        &1.000       &{\bf 0.963}\\
{\bf 0.818}      &-0.366        &{\bf 0.963}       &1.000
\end{bmatrix}\eeq
\vspace{-.3em}

\begin{figure}[]
    \centering
  \includegraphics[width=.482\linewidth,trim=.5cm .5cm .5cm .5cm]{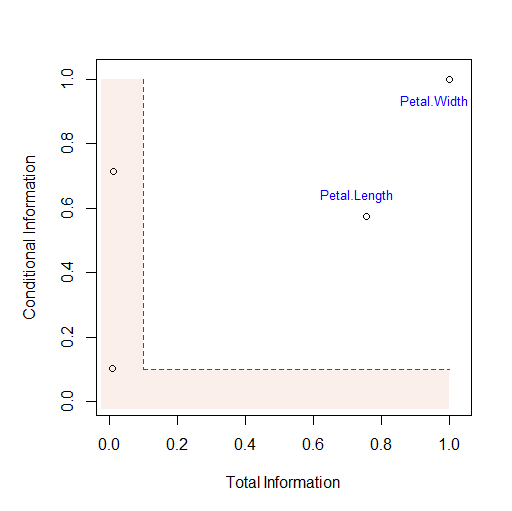}~~~~~
   \includegraphics[width=.5\linewidth,trim=.5cm .5cm .5cm .5cm]{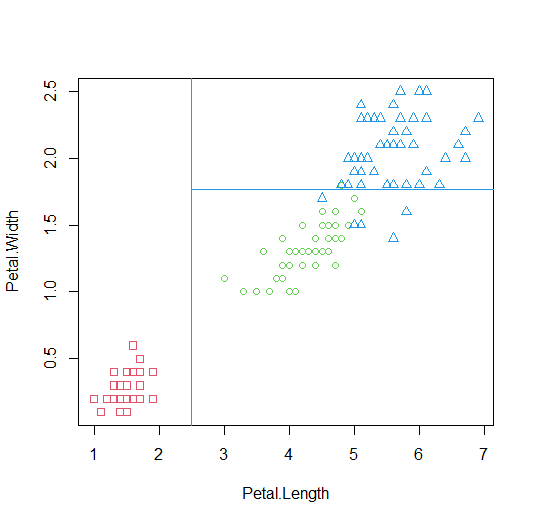}\\[2em]
  \includegraphics[width=.7\linewidth,trim=.5cm .5cm .5cm .5cm]{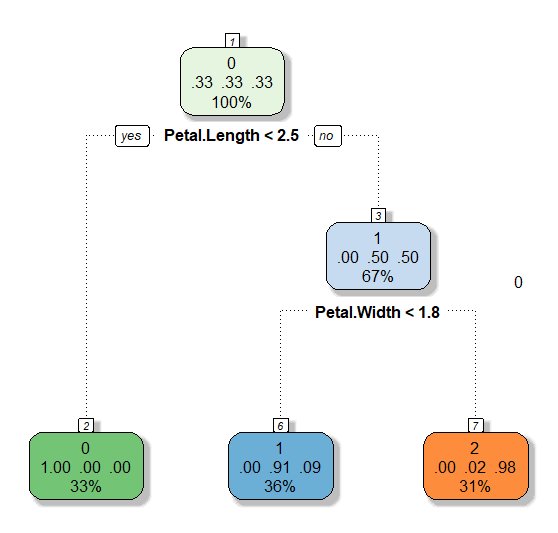}
  \vskip2em
    \caption{Iris data analysis. Top left: infogram; top right: the scatter plot of the data based on the selected core features; three different classes are indicated by red, green, and blue colors; bottom: the estimated decision tree classifier using the variables petal-length and petal-width.} \label{fig:iris}
  \end{figure}

The infogram for the iris data, constructed using the recipe given in section \ref{sec:AINTER}, is shown at the top-left corner of Fig. \ref{fig:iris}, which clearly identifies \texttt{petal.length} and \texttt{petal.width} as the \textit{core} relevant features. Since we have reduced the problem to a bivariate one (variables: \texttt{petal.length} and \texttt{petal.width}), we can now simply plot the data. This is done in the top-right of Fig. \ref{fig:iris}. We can even visually draw the linear decision surfaces to separate the three classes; see the red and blue lines in the scatter plot. Finally, we train a decision tree classifier based on the selected core features: \texttt{petal.length} and \texttt{petal.width}. The estimated \texttt{COREtree} is shown in the bottom panel, which gives a beautifully crisp (readily interpretable) decision rule for classifying iris flowers.

 \vspace{.3em}
\subsection{EU’s Artificial Intelligence Act }
\label{sec:EUAI}
On 21st April 2021, the European Union (EU) unveiled strict regulations\footnote{The full report is available online at https://bit.ly/EUAI\_act. Also see the  New York Times article https://www.nytimes.com/2021/04/16/business/artificial-intelligence-regulation.html
} to govern high-risk AI systems, which provides one of the first formal and comprehensive regulatory frameworks on AI. Few key takeaways from the report:

~~$\bullet$ A risk management system shall be established, implemented, documented and maintained in relation to high-risk AI systems.

~~$\bullet$ In identifying the most appropriate risk management measures, the following shall be ensured: elimination or reduction of risks as far as possible through adequate design and development.

~~$\bullet$  Bias monitoring, detection, and correction mechanism should be at place for high-risk AI systems in the pre-as well as the post-deployment stages.

~~$\bullet$ High-risk AI systems shall be designed and developed in such a way to ensure that their operation is sufficiently transparent to enable users to interpret the system’s output and use it appropriately.

~~$\bullet$  High-risk AI systems should equip with appropriate human-machine interface tools---which allows the system to be effectively overseen by natural persons during the period in which the AI system is in use.

~~$\bullet$ High-risk AI technology providers shall ensure that their systems undergo 
regulatory compliant assessments. If the AI system is not in conformity with the requirements, they need to take the necessary corrective actions before putting them into service. Companies that fail to do so could face fines of up to $6\%$ of their global sales.

\end{document}